\documentclass[lettersize,journal]{IEEEtran}
\usepackage{amsmath,amsfonts}
\usepackage{array}
\usepackage{textcomp}
\usepackage{stfloats}
\usepackage{url}
\usepackage{verbatim}
\usepackage{graphicx}
\usepackage{cite}
\usepackage{xcolor}
\usepackage{verbatim}
\usepackage{subcaption}
\usepackage{multirow}

\usepackage{algorithmicx}
\usepackage[ruled,lined, noend,linesnumbered]{algorithm2e}

\graphicspath{{./}}

\usepackage{hyperref}

\newcommand{\orcid}[1]{\textsuperscript{\href{https://orcid.org/#1}{\includegraphics[width=8pt]{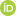}}}}

\hypersetup{
    colorlinks=true,       
    linkcolor=black,        
    citecolor=blue,        
    urlcolor=black,         
    pdfborder={0 0 0}      
}

\begin{document}

\title{\texttt{OmniLearn}: A Framework for Distributed Deep Learning over Heterogeneous Clusters}

\author{Sahil Tyagi \orcid{0009-0007-8314-4745},
  Prateek Sharma \orcid{0000-0003-1789-0145}, \textit{Member, IEEE}
\thanks{This is an extended version of the preliminary paper: Tyagi, Sahil, and Sharma, Prateek. ``Taming Resource Heterogeneity in Distributed ML Training with Dynamic Batching''. In 2020 IEEE International Conference on Autonomic Computing and Self-Organizing Systems (ACSOS) \cite{bb50}.}
}

\markboth{IEEE Transactions on Parallel and Distributed Systems}%
{Shell \MakeLowercase{\textit{et al.}}: A Sample Article Using IEEEtran.cls for IEEE Journals}

\maketitle

\begin{abstract}
	Deep learning systems are optimized for clusters with homogeneous resources.
However, heterogeneity is prevalent in computing infrastructure across edge, cloud and HPC.
When training neural networks using stochastic gradient descent techniques on heterogeneous resources, performance degrades due to stragglers and stale updates. 
In this work, we develop an adaptive batch-scaling framework called \texttt{OmniLearn} to mitigate the effects of heterogeneity in distributed training. 
Our approach is inspired by proportional controllers to balance computation across heterogeneous servers, and works under varying resource availability.
By dynamically adjusting worker mini-batches at runtime, 
\texttt{OmniLearn} reduces training time by 14-85\%.
We also investigate asynchronous training, where our techniques improve accuracy by up to 6.9\%. 
\end{abstract}


\section{Introduction}

Distributed training of large machine learning models is a major workload in cloud and high-performance computing (HPC) clusters. 
Models are trained by running thousands of iterations of stochastic gradient descent (SGD), where the gradients computed by each worker (i.e., a server) are periodically collectively communicated and aggregated.

The SGD theoretical framework and distributed training systems assume that all workers are ``homogeneous'', such that all workers compute gradients at the same speed.
However, virtual clusters in data centers and especially clouds \emph{do not} always exhibit this resource homogeneity. 
The performance of different workers can be affected due to performance interference with co-located applications; workers may be throttled by the cloud or data center provider; or the cluster may have servers with vastly different resource configurations.

This \emph{resource heterogeneity} is a key characteristic of cloud-based applications, and distributed ML model training must be able to tolerate and perform well even in heterogeneous environments. 
However, heterogeneity presents many fundamental challenges to distributed training: Bulk Sychronous Parallel (BSP) leads to  stragglers, which reduces the parallel efficiency and thus increases the total training time.

While \emph{asynchronous} model training can alleviate such stragglers, in practice, the stale gradients leads to poor statistical efficiency~\cite{b18}, and low model quality (i.e., generalizability). 
Thus, heterogeneity affects both the dominant parallelism paradigms, and is a \emph{fundamental} problem in distributed machine learning. 

Our primary insight is that the performance degradation due to heterogeneity can be alleviated through \emph{dynamic mini-batch sizing}.
Instead of giving all workers fixed-sized mini-batches for gradient computation, we adjust their sizes carefully based on the throughput of each worker.
We use proportional control where each worker adjusts its batch size based on throughput and staleness~\cite{b2}. 
This significantly reduces stragglers and gradient staleness, and we show it to be a general technique suitable in both  synchronous and asynchronous training.

We develop and implement these batch size scaling techniques as part of \texttt{OmniLearn}, a machine learning framework for model training on heterogeneous resources \footnote{Source code available at \href{https://github.com/sahiltyagi4/OmniLearn}{https://github.com/sahiltyagi4/OmniLearn}.}.
Unlike prior work \cite{bb42, bb41} which regards heterogeneity as a result of stochastic performance variations, we focus on the systemic and ``static'' heterogeneity where the workers may have vastly different resource configurations (such as the number of CPU cores and memory).
This allows us to efficiently train on clusters where the ratio of fastest to slowest worker may be as high as $8\times$.

In addition to the large static heterogeneity, our continous batch size scaling also helps alleviate \emph{dynamic} heterogeneity which can arise due to performance interference in shared clusters or from concurrent jobs. 
In summary, we make the following contributions: 
\begin{itemize}
\item \texttt{OmniLearn} is the first solution for heterogeneous training in both BSP and ASP paradigms. In BSP it eliminates stragglers and improves parallel efficiency. In ASP, it  minimizes staleness in gradient updates and improves statistical efficiency.
\item Implement a zero-configuration, black-box framework over PyTorch and TensorFlow that accelerates training over heterogeneous resources, i.e., nodes with non-uniform or elastic resource configurations. 
\item We empirically evaluate various neural networks in settings with \textit{static} and \textit{dynamic} heterogeneity to assess the overall compute cost, training time and convergence quality, and compare with `uniform-batching' (i.e., data-parallel training with identical batch-size across workers).
\end{itemize}


\section{Background And Motivation}

In this section, we discuss the parallel and statistical performance trade-offs in synchronous and asynchronous training. We also focus on how resource heterogeneity affects each, and the related work in the area. 

\subsection{Distributed Data-Parallel Training}\label{subsec:distdptrainBGRelated}

Model training is an iterative-convergent process that updates a model's parameters (i.e., weights) at each step or iteration, and minimizes loss via optimization techniques like gradient descent \cite{b3}.
BSP achieves high throughput via weak-scaling: same batch-size on each worker so $K$ times more samples are processed in each training step as the cluster is scaled from 1 to $K$ workers. 
A worker $k$ computes updates on a mini-batch $\mathit{b_{(i, k)}}$ of size $|\mathit{b}|$ from dataset $\mathcal{B}_{k}$ to update model parameters `$\theta$' at step $(i+1)$.
Equation~(\ref{eqn:distsgdBSP}) shows the mini-batch gradient descent update where gradients computed on loss function $\mathcal{F}(\cdot)$ are scaled by learning-rate $\eta$ and averaged together before applying them:

\begin{subequations}
  \begin{equation}
    \theta_{i+1} = \theta_{i} - \eta \dfrac{1}{K} \sum_{k=1}^{k=K} \dfrac{\partial}{\partial \theta_{i}} \left(\dfrac{1}{|b|} \sum_{\mathit{b_{(i, k)}} \in \mathcal{B}_{k}} \mathcal{F}(b_{(i, k)},\theta_{i})\right)
    \label{eqn:distsgdBSP}
  \end{equation}
  \begin{equation}
    t_{iteration} = t_{compute} + t_{sync} + t_{data {\text -}IO}
    \label{eqn:bspitrtime}
  \end{equation}
\end{subequations}

The synchronization step is \textit{blocking}, so training is halted from proceeding to the next step until each worker calculates and communicates its updates across the cluster.
The iteration time in BSP is thus comprised of computation cost, update synchronization time, and I/O overheads (Equation~(\ref{eqn:bspitrtime})).
The term $t_{data {\text -}IO}$ is the cumulative overhead of reading and batching training samples, data-movement overhead, etc.
Based on cluster topology, BSP may be centralized using a parameter server (PS) or communicate via MPI collectives in decentralized systems, each with its distinct communication costs~\cite{b4, b37}.
Frameworks like PyTorch RPC~\cite{bb78} and BytePS~\cite{b8} implement centralized training, while libraries like PyTorch DDP \cite{bb78} perform decentralized training. 

With a centralized PS, ASP involves each worker training independently and sending its updates asynchronously to update the globally shared model maintained by the PS.
Each worker computes its gradient updates locally and pushes the updated parameters to the PS in a lock-free and non-blocking manner, then proceeding to the next iteration as shown in Equation~(\ref{eqn:aspsgd}).
Compared to BSP, communication overhead in ASP is low, since  there is no synchronization barrier between workers.
Apart from centralized PS, there are decentralized variants of ASP as well.
Gossip-based GoSGD \cite{b11} and AD-SGD \cite{b12} involve random walks over the cluster and gradient exchange between a subset of nodes chosen randomly at each iteration.
But without periodic cluster-wide synchronization, local model replicas may diverge considerably, while infrequent global aggregation may degrade a model's convergence quality.


\begin{equation}
  \theta_{i+1}^{(k)} = \theta_{i}^{(k)} - \eta \dfrac{\partial}{\partial \theta_{i}^{(k)}}(\dfrac{1}{|b|} \sum_{\mathit{b}_{(i,k) \in \mathcal{B}_{k}}} \mathcal{F}(\mathit{b}_{(i,k)}, \theta_{i}^{(k)})) \;\forall\; k \in [1, K]
  \label{eqn:aspsgd}
\end{equation}

\vspace{-1em}

\subsection{Systems Heterogeneity in Distributed Training}\label{subsec:heterogeneityDDP}

Since resource heterogeneity is ubiquitous from edge to cloud, applications are often deployed on nodes with varying sizes and compute resources.
Thus, it is crucial to train efficiently over heterogeneous devices on the edge, as well as reduce training time over low-end servers in datacenters~\cite{b13}.
Distributed training must be ``omnivorous'' and able to leverage a variety of compute resources, rather than assume homogeneous environments with \textit{static} resource availability.
Further, modern applications must also contend with \textit{dynamic} heterogeneity from varying resource availability arising from concurrent jobs in shared clusters.
Burstable cloud VMs \cite{bb1314, b14} are also dynamically heterogeneous as they can burst at different times.
Finally, federated methods~\cite{b7, b10, bb39} either process samples according to data-distribution or dataflow-rates on each device, so devices with more data process more samples during training. In such cases, compute heterogeneity arises from extrinsic factors rather than due to intrinsic aspect of computational capability of a device.

Heterogeneity impacts many aspects of gradient descent performance, in both synchronous and asynchronous settings. 
BSP suffers from parallel inefficiency, while ASP struggles with poor statistical performance with low convergence quality. 
By design, BSP has a synchronization barrier where training processes halt until updates are aggregated and circulated back to every node; heterogeneity thus can thus be detrimental due to \emph{straggler slowdown} as nodes with lower computational capability take longer to process a mini-batch than those with more resources.
After computing gradients, fast nodes sit idle and waste cycles while waiting for slow nodes to complete their local updates.
From Equation~(\ref{eqn:bspitrtime}), higher compute cost along with communication overhead affects parallel performance \cite{b16, b17}, thus raising iteration costs and consequently, overall training time.

Stragglers do not affect parallel performance in ASP training as there is no wait-period on slower nodes, but do impact the convergence quality (i.e., model generalization) due stale updates.
This is because the progress made by faster nodes is overridden by slow ones that compute their updates from a much older model replica hosted locally \cite{b18}.
Simply training a model for longer may not necessarily yield desired gains in model accuracy.
As we show in \S \ref{sec:omnilearnintro}, models training with ASP may not achieve BSP-level accuracy despite running for significantly longer iterations.

Methods like DynSGD and ConSGD \cite{bb41} mitigate staleness or stragglers by using a vector-clock technique and dynamic learning-rate, while Hop~\cite{bb42} performs decentralized training uses a bounded staleness approach to limit the iteration-gap.
\cite{bb46, bb44, bb43}  mitigates stragglers in BSP via optimized partial collective communication, while  \cite{bb45} adds backup workers that completely drop slower nodes in distributed training.
Petrel \cite{bb40} segregates nodes into fast and slow groups, and performs hybrid communication over these disjoint communities.
These works view heterogeneity from the perspective of random worker slowdowns instead of one that is inherent in a cluster due to differently sized clusters.
Additionally, this stochastic performance variability approach does not necessarily reflect worker slowdowns due to performance interference from concurrent jobs in shared clusters, or burstable or transient resources like spot VMs; a form of dynamic heterogeneity we address in \texttt{OmniLearn}.

There are additional optimizations on top of ASP/BSP such as communication scheduling and gradient bucketing \cite{bb48, b37} in BSP which improves communication cost, or stale-synchronous training \cite{bb47, b10} over ASP which mitigates staleness by imposing a staleness-threshold between fast and slow workers.
These are not designed for systemic high heterogeneity and thus complementary to our approach.

\BlankLine
\noindent \textit{\textbf{Summary:} Training over heterogeneous clusters with communication patterns like ASP and BSP introduces low-efficient states like staleness and stragglers.
Staleness degrades statistical performance of ASP training such that the progress made by fast workers is overwritten by stale updates from slow workers.
In BSP, stragglers cause slowdown on fast workers which must wait for updates from slow workers at the synchronization barrier.
We assess the impact of stragglers and staleness on model performance in \S\ref{subsec:designHeteroImpact}.}

\begin{table}
  \renewcommand{\arraystretch}{1.3}
  \caption{Heterogeneity configurations (CPU-core allocation) across 4 workers.}
  \centering
  \begin{tabular}{|c|c|c|c|c|c|}
    \hline
    \bfseries HL & \bfseries Method & \bfseries Node \#1 & \bfseries Node \#2 & \bfseries Node \#3 & \bfseries Node \#4 \\
    \hline
    \multirow{2}{*}{\textit{HL1}} & BSP & 12 & 12 & 12 &12 \\
    \cline{2-6}
                 & ASP & 10 & 10 & 10 & 10 \\
    \hline
    \multirow{2}{*}{\textit{HL2}} & BSP & 12 & 12 & 8 & 16 \\
    \cline{2-6}
                 & ASP & 8 & 8 & 8 & 16 \\
    \hline
    \multirow{2}{*}{\textit{HL4}} & BSP & 9 & 9 & 6 & 24 \\
    \cline{2-6}
                 & ASP & 10 & 10 & 4 & 16 \\
    \hline
    \multirow{2}{*}{\textit{HL8}} & BSP & 6 & 6 & 4 & 32 \\
    \cline{2-6}
                 & ASP & 4 & 4 & 4 & 28 \\
    \hline
  \end{tabular}
  \label{table:computecoreHLs}
\end{table}

\subsection{Measuring Heterogeneity and its Impact on Training}\label{subsec:designHeteroImpact}

\begin{figure}
  \hspace{-0.25cm}
  \begin{subfigure}{0.22\textwidth}
    \includegraphics[width=1.1\linewidth]{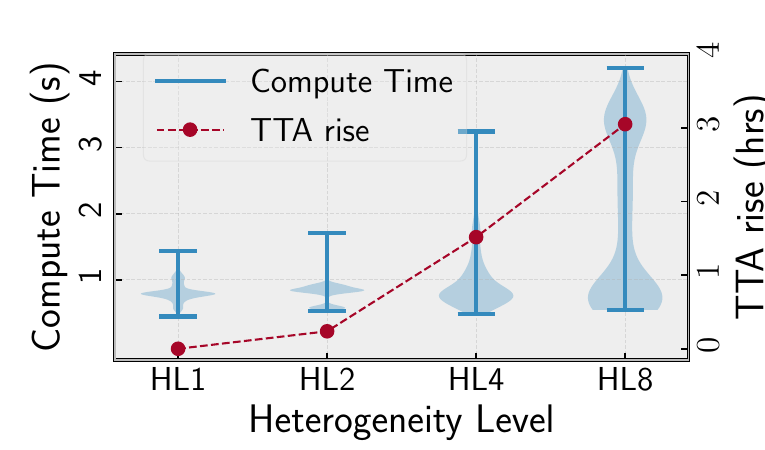}
    \caption{ResNet18}
    \label{fig:bspbaselineResNet18}
  \end{subfigure}
  \hspace{1em}
  \begin{subfigure}{0.22\textwidth}
    \centering
    \includegraphics[width=1.1\linewidth]{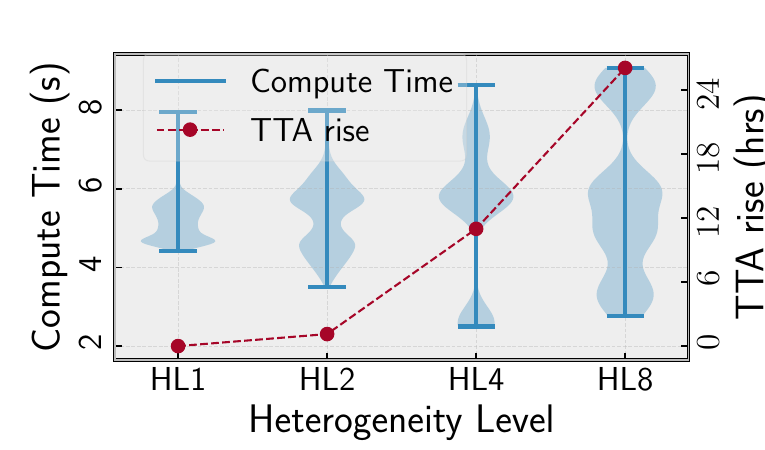}
    \caption{ResNet50}
    \label{fig:bspbaselineResNet50}
  \end{subfigure}
  
  \vspace{0.25em}
  
  \hspace{-0.25cm}
  \begin{subfigure}{0.22\textwidth}
    \includegraphics[width=1.1\linewidth]{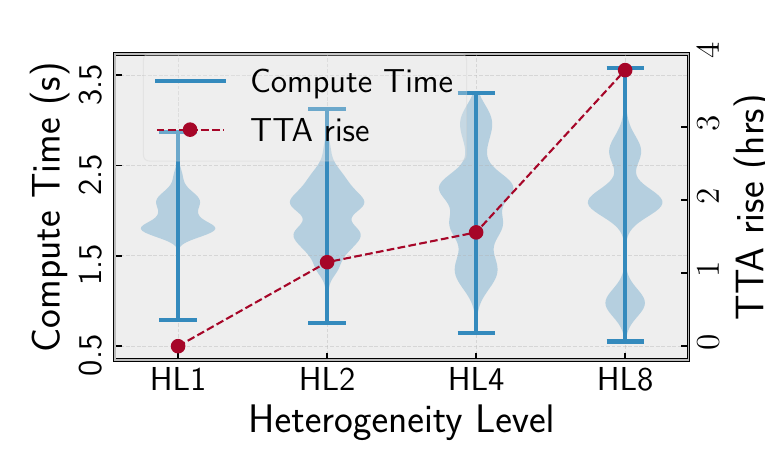}
    \caption{AlexNet}
    \label{fig:bspbaselineAlexNet}
  \end{subfigure}
  \hspace{1em}
  \begin{subfigure}{0.22\textwidth}
    \centering
    \includegraphics[width=1.1\linewidth]{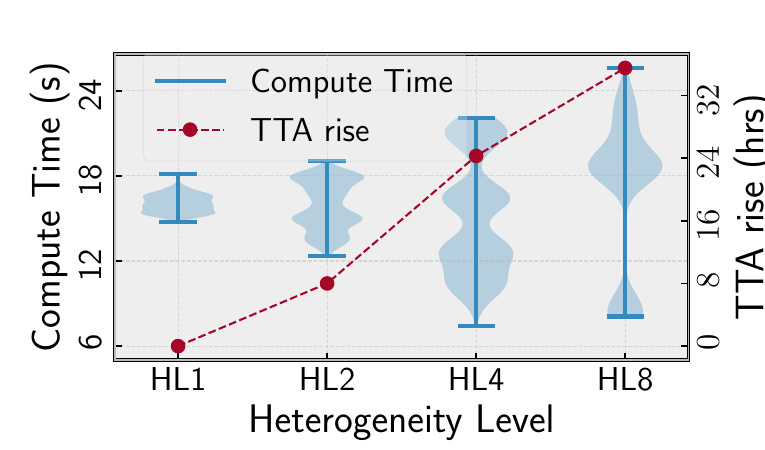}
    \caption{VGG11}
    \label{fig:bspbaselineVGG}
  \end{subfigure}
  \caption{BSP worker compute-time distribution across \textit{HL}s and rise in time-to-accuracy (TTA) relative to \textit{HL1} (in hours).}
  \label{fig:bsdptimedistbaseline}
\end{figure}

We now empirically examine and quantify the effect of heterogeneity on training performance. 
To exclusively study the impact of different degrees of heterogeneity, we keep the cumulative computational resources in a cluster the same across various cluster configurations.
For a cluster of CPU-based workers, we measure heterogeneity by \emph{Heterogeneity-level (HL)} that compares the total cores available on the largest worker to that on the smallest worker (from Equation~(\ref{eqn:hlevel})).
For e.g., a 4-worker cluster with CPU-cores (6,6,4,24) respectively has \textit{HL} 6.0, while a homogeneous cluster with (10,10,10,10) cores has \textit{HL} 1.0.
Although both clusters have the same 40 cores in total, the distribution of compute resources varies across the workers in the two cases.
Heterogeneity in ML accelerators like GPUs or TPUs can be measured by comparing the total floating-point operations per-second (FLOPs), compute cores available (like CUDA or Tensor cores) or available worker memory.
Metrics like \textit{HL} based on min-max are commonly used to emulate heterogeneity in BSP and ASP communication models as they can effectively capture the straggler/staleness relationship between compute nodes.

\begin{equation}
  \textit{Heterogeneity-level (HL)} = \dfrac{\text{max.} \; \text{\#} \; \text{cores or FLOPs}}{\text{min.} \; \text{\#} \; \text{cores or FLOPs}}
  \label{eqn:hlevel}
\end{equation}

Equation~(\ref{eqn:hlevel}) measures the computational variance between workers in an offline manner, i.e., without actually running the model to measure performance.
We evaluate heterogeneity over a 4-node cluster via docker and limit CPU cores across each worker (details in \S \ref{subsec:clustermodeldescribe}).
From Table~(\ref{table:computecoreHLs}), in \textit{HL1} each worker comprises of 12 CPU-cores in BSP, and 10 cores in ASP.
The latter contains an additional 8-core PS node so cumulative cores in both centralized ASP and decentralized BSP is the same (= 48).
Under these constraints, the largest \textit{HL} ASP allocation attains is 7.0, but we approximate it to \textit{HL8} since its still a highly heterogeneous configuration and the purpose of \textit{HL} metric is only to emulate varying degrees of heterogeneity.
We increase heterogeneity by widening the divergence between workers \#3 and \#4.
For e.g., BSP workers at \textit{HL8} contain (6, 6, 4, 32) CPU-cores so $32/4$ = 8.0.
Across all \textit{HL}s, the cumulative core-count is also fixed to 48-cores.
We evaluate \emph{heterogeneity} in the context of ASP and BSP by training over various configurations from Table~(\ref{table:computecoreHLs}).

In \textit{uniform-batching}, we train various models over different \textit{HL}s while keeping per-worker batch-size fixed.
Due to the synchronous nature of BSP, the convergence quality will be the same across all \textit{HL}s in uniform-batching as statistical performance stays the same.
However, parallel efficiency suffers due to stragglers, raising the total iteration cost (and thus, training time).
Figure~(\ref{fig:bsdptimedistbaseline}) shows the worker compute times in BSP for various \textit{HL} configurations from Table~(\ref{table:computecoreHLs}).
ResNet18, ResNet50 \cite{b26}, AlexNet \cite{b28} and VGG11 \cite{b31} converge to \textit{95\%}, \textit{88\%}, \textit{70\%} and \textit{83\%} test accuracy after training for \textit{30K}, \textit{80K}, \textit{15K} and \textit{15K} iterations respectively.
It is important to note that \emph{only} the parallel performance (not statistical) changes with heterogeneity in BSP uniform-batching.
We also report the additional increase in training time to reach the desired accuracy over a homogeneous cluster (i.e., \textit{HL1}) in Figure~(\ref{fig:bsdptimedistbaseline}).
Workers with lower compute resources take more time to calculate gradients over the same mini-batch size.
As heterogeneity gets worse when \textit{HL} rises, divergence in worker computation time increases as well, leading to stragglers and longer iteration times.
The density distribution is thus more spread apart (i.e., first and third quartiles) when \textit{HL}s rise, and the median compute time (second quartile) is higher.
\emph{Due to stragglers, training time in high heterogeneity settings can increase between 3 to 35 hours over a homogeneous cluster.}

\begin{table}
  \centering
  \caption{ASP training at different \textit{HL}s}
  \begin{tabular}{|c|c|c|c|c|}
    \hline
    \bfseries Model & \bfseries Iterations & \bfseries HL & \bfseries Accuracy & \bfseries TTA (hrs)\\
    \hline
    \multirow{3}{*}{ResNet18} & \multirow{3}{*}{$120 \times 10^{3}$} & \textit{HL1} & 89.16\% & 22.2\\
    \cline{3-5}
                    & & \textit{HL4} & 86.24\% & 15.5\\
    \cline{3-5}
                    & & \textit{HL8} & 77.68\% & 12.9\\
    \hline
    \multirow{3}{*}{ResNet50} & \multirow{3}{*}{$300 \times 10^{3}$} & \textit{HL1} & 82.64\% & 150.4 \\
    \cline{3-5}
                    & & \textit{HL4} & 75.93\% & 145.3\\
    \cline{3-5}
                    & & \textit{HL8} & 49.51\% & 116.2\\
    \hline
    \multirow{3}{*}{AlexNet} & \multirow{3}{*}{$30 \times 10^{3}$} & \textit{HL1} & 65.48\% & 8.1 \\
    \cline{3-5}
                    & & \textit{HL4} & 58.86\% & 7.9\\
    \cline{3-5}
                    & & \textit{HL8} & 54.32\% & 6.2\\
    \hline
    \multirow{3}{*}{VGG11} & \multirow{3}{*}{$48 \times 10^{3}$} & \textit{HL1} & 82\% & 107.3 \\
    \cline{3-5}
                    & & \textit{HL4} & 77.16\% & 114.4\\
    \cline{3-5}
                    & & \textit{HL8} & 68.24\% & 81.9\\
    \hline
  \end{tabular}
  \label{table:aspbaselineAccTTA}
\end{table}

\begin{figure}
  \hspace{-0.2cm}
  \begin{subfigure}{0.22\textwidth}
    \includegraphics[width=1.1\linewidth]{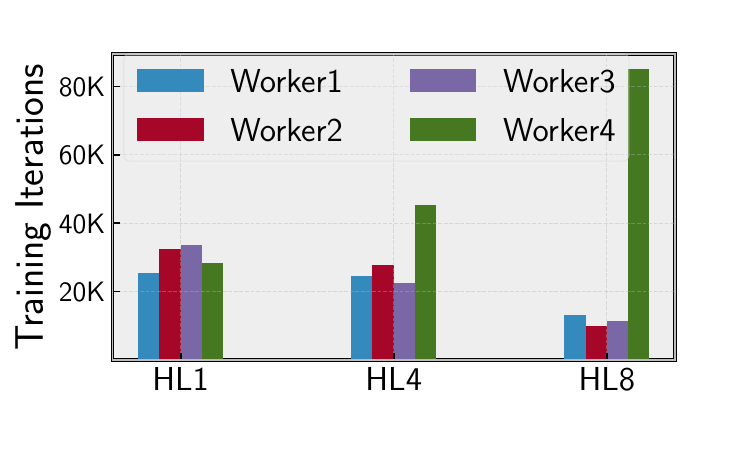}
    \caption{ResNet18}
    \label{fig:aspbaselineResnet18itrdensity}
  \end{subfigure}
  \hspace{1.5em}
  \begin{subfigure}{0.22\textwidth}
    \centering
    \includegraphics[width=1.1\linewidth]{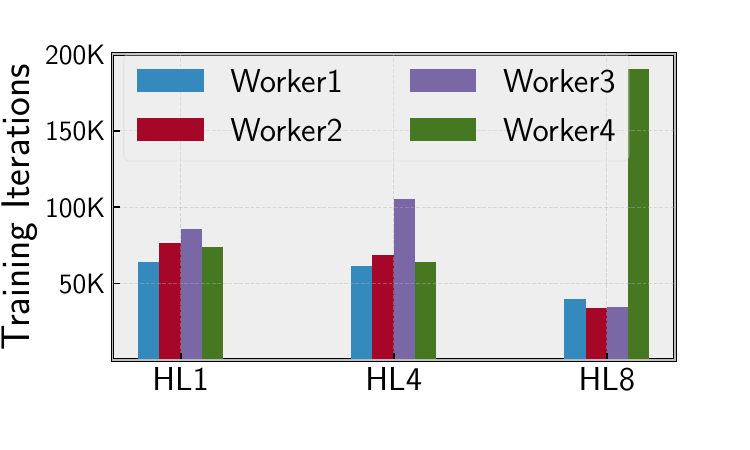}
    \caption{ResNet50}
    \label{fig:aspbaselineResnet50itrdensity}
  \end{subfigure}
  
  \vspace{0.25em}
  
  \hspace{-0.2cm}
  \begin{subfigure}{0.22\textwidth}
    \includegraphics[width=1.1\linewidth]{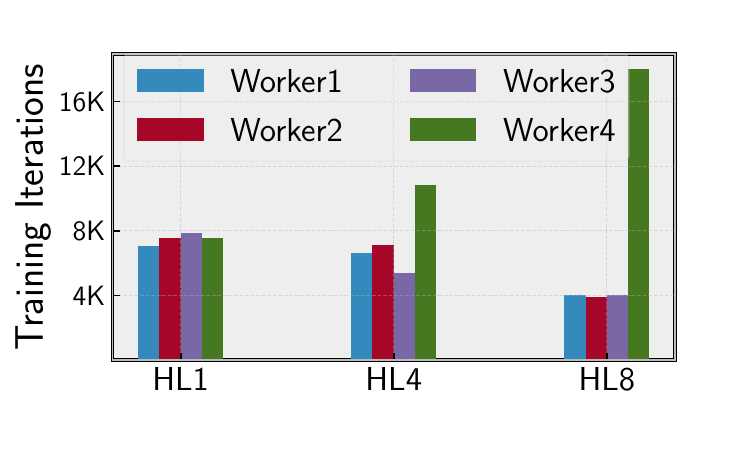}
    \caption{AlexNet}
    \label{fig:aspbaselineAlexnetitrdensity}
  \end{subfigure}
  \hspace{1.5em}
  \begin{subfigure}{0.22\textwidth}
    \centering
    \includegraphics[width=1.1\linewidth]{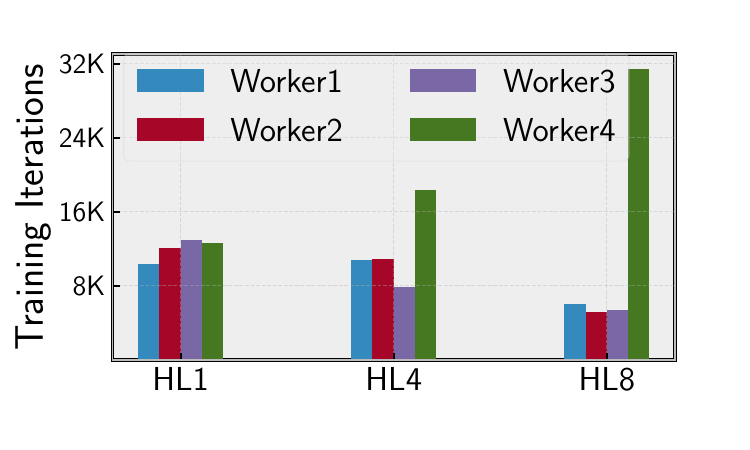}
    \caption{VGG11}
    \label{fig:aspbaselineVGGitrdensity}
  \end{subfigure}
  \caption{ASP's iteration distribution of model updates from different workers across \textit{HL}s.
    The accuracy degradation at high \textit{HL}s is attributed to staleness from skewed distribution of training iterations.}
  \label{fig:aspbaselineplotsitrDense}
\end{figure}

In case of ASP, parallel efficiency is improved at the detriment of statistical performance.
The accuracy degradation, total iterations taken and time-to-accuracy (TTA) under ASP is shown in Table~(\ref{table:aspbaselineAccTTA}).
Training time decreases as more resources are added to worker \#4 at high heterogeneity levels.
Another interesting aspect of ASP is that it takes considerably more iterations to converge than BSP.
For instance, ResNet18 takes \textit{30K} iterations to converge in BSP, and over \textit{120K} in case of ASP.
On the other hand, the statistical performance of ASP is significantly worse than BSP, as seen from its lower accuracy levels \textit{even at the same HL}.
For e.g., ResNet50 at \textit{HL1} reaches only 82.64\% test accuracy with ASP and 88\% with BSP.
Within ASP itself, convergence quality further degrades as heterogeneity rises.
This is attributed to staleness, which gets worse as \textit{HL} increases and slower workers take longer to compute updates, which may override the latest model state from faster workers.
We confirm this in Figure~(\ref{fig:aspbaselineplotsitrDense}) which shows the frequency distribution of model updates performed on the central PS by each worker.
A worker with more compute resources will have a high frequency than slower nodes.
A homogeneous cluster (\textit{HL1}) with uniform resources has all workers equipped with similar compute and thus, PS receives uniform volume of updates from every node.
As heterogeneity rises, iteration distribution skews towards workers with more resources as they tend to make considerably more updates than slower ones.
As per Table~(\ref{table:computecoreHLs}), worker \#4 gets more resources allocated to it and so has a higher iteration frequency in heterogeneous settings like \textit{HL4} and \textit{HL8}.
\section{\large{\texttt{OmniLearn}}\: design}\label{sec:omnilearnintro}

\begin{figure}[t]
  \hspace{-0.72cm}
  \includegraphics[width=1.15\linewidth]{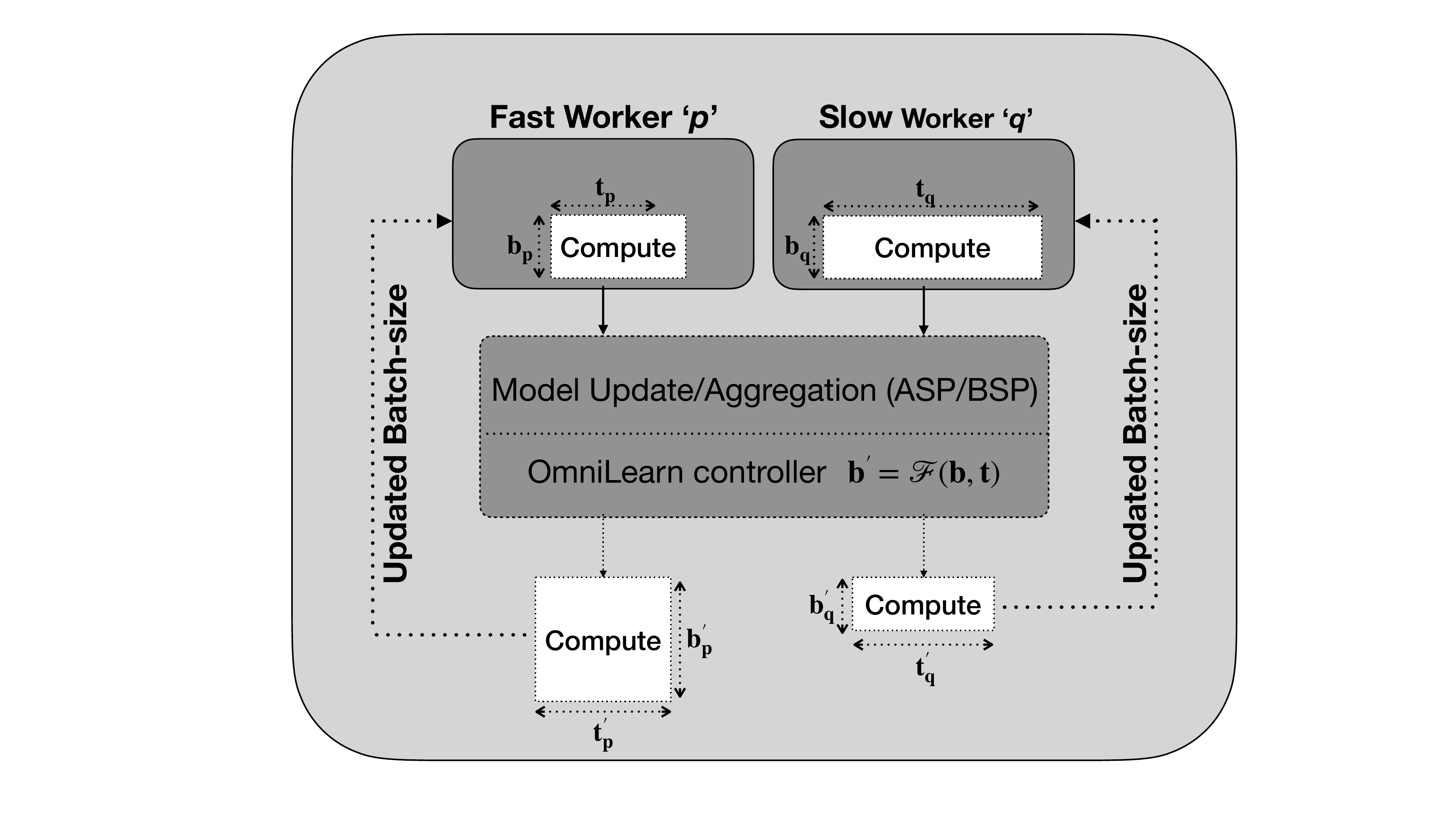}
  \caption{\texttt{OmniLearn} overview:
    Initially, fast and slow workers ($p$, $q$) train on the same mini-batch $b_{p} = b_{q}$ (breadth of `Compute' block) that leads to stragglers (BSP) or staleness (ASP) as compute times $t_{p} < t_{q}$ (length of `Compute').
Controller adjusts mini-batches to equalize compute times, i.e., $t^{'}_{p} \sim t^{'}_{q} \; : \; b^{'}_{p} > b^{'}_{q}$ since $p > q$ from computational standpoint.}
  \label{fig:omnidesignfig}
\end{figure}

Synchronous and asynchronous mechanisms offer different trade-offs in distributed deep learning, and using one over the other depends on a ML practitioner's primary objective of attaining maximal speedup vs. best-accuracy.
In heterogeneous clusters, BSP suffers from wait-time at the synchronization barrier.
As demonstrated in \S \ref{subsec:designHeteroImpact}, ASP accelerates each training step by avoiding stragglers, but needs many more iterations to converge and generally attains lower model quality than BSP.
This is due to staleness in updates computed from outdated model replicas.
These training mechanisms suffer from different challenges due to heterogeneity.


\noindent \textbf{Key Design Principle.}
As heterogeneity is pervasive and nodes in a cluster have varying available resources, stragglers or staleness may be mitigated by ensuring different workers have roughly the same computation cost or time.
Thus, the basic principle we will use is equalization of work across workers, with per-worker dynamic batch-size adjustment as the key mechanism.
If $t_{compute}$ (from Equation (\ref{eqn:bspitrtime})) is similar across workers in BSP, fast workers do not have to wait for the slow workers to complete their steps, thus eliminating stragglers altogether.
Similarly, if compute cost across ASP workers is the same, the local and global model states (on workers and PS respectively) would not diverge, and no updates from outdated an model will be applied on the PS.
\emph{The key insight of \texttt{OmniLearn} is to perform variable amount of work (i.e., worker's batch-size) on each node based on its resource availability or compute capability}, as illustrated in Figure~(\ref{fig:omnidesignfig}).
Unlike prior work which seeks to address BSP stragglers and ASP staleness through different mechanisms (such as vector-clocks and learning-rate adjustment described in the previous section), we propose a single generalizable technique which addresses both.

For the chosen communication pattern, computation time is registered across each worker corresponding to its local batch-size.
Based on either the \textit{HL} metric or proportional control, we vary the amount of work $b$ on each worker (i.e., its local batch-size shown as the breadth of `Compute' block in Figure~(\ref{fig:omnidesignfig})) in order to equalize their respective computation times $t$ (shown as the length of `Compute' block).
Using \textit{HL} metric under static heterogeneity, we develop  \emph{variable-batching} based on static resource allocation, i.e., workers are allotted batch-size based on available computational resources.
However, such static policies fail to accurately partition batches which equalize computation time across all workers.
For a more accurate batch-estimation, we develop a proportional control-based \emph{dynamic-batching} for precise allocation under both static and dynamic heterogeneity.
The ``controller'' component in Figure~(\ref{fig:omnidesignfig}) differs for each one, as explained in \S \ref{subsec:bspvarbatch} and \ref{subsec:propctrldynhet}.
As for the differences in PID control mechanism for BSP and ASP, we describe the controller for each in Algorithm (1) and (2) respectively.

\subsection{Variable-Batching for Static Heterogeneity}\label{subsec:bspvarbatch}

Intuitively, batch-sizes need not be identical across workers.
Instead, mini-batches should be proportional to a worker's resource availability.
Thus, \texttt{OmniLearn} assigns varying amounts of work to perform on different nodes, in order to minimize the divergence in worker compute times.
The ``controller'' component in Figure~(\ref{fig:omnidesignfig}) for \textit{variable-batching} implements Equations~(\ref{eqn:BSPVarBatch}) and (\ref{eqn:ASPVarBatch}) for BSP and ASP respectively, further explained below.

\subsubsection{Synchronous Training}

In BSP, the global batch-size is fixed as it is a crucial training hyperparmeter.
Assuming per-worker initial batch-size `$b$', the global batch-size processed in one iteration over a cluster of `$K$' workers is $\sum_{i=1}^{K}b_{i} = Kb$.
In \textit{variable-batching}, we reduce the batch-size on slower nodes while proportionally raising it on faster workers, illustrated with the following equations: 

\begin{subequations}
  \begin{equation}
    b_{k} = \dfrac{c_{k}}{\sum_{k=1}^{K}c_{k}} \cdot Kb
    \label{eqn:varbszBSP}
  \end{equation}
  \begin{equation}
    g_{(k, i)} = \mathbf{\lambda_{k}} \odot \dfrac{\partial}{\partial \theta_{i}}(\dfrac{1}{|b_{k}|} \sum_{b_{(i, k)} \in \mathcal{B}_{k}} \mathcal{F}(b_{(i, k)}, \theta_{i})) \Rightarrow \mathbf{\lambda_{k}} = \frac{b_{k}}{\sum_{i=1}^{K}b_{i}}
    \label{eqn:bspweightedgrads}
  \end{equation}
  \begin{equation}
    \theta_{i+1} = \theta_{i} - \eta \odot \sum_{k=1}^{K} g_{(k, i)}
    \label{eqn:bspvariableSGD}
  \end{equation}
  \label{eqn:BSPVarBatch}
\end{subequations}

In Equation~(\ref{eqn:varbszBSP}), $c_{k}$ is the total compute resource available on worker $k$, while the denominator corresponds to cumulative cores available across the cluster.
Gradients computed on workers training with larger batches are more accurate and representative of the true gradients compared to those calculated with a small batch-size \cite{b19, b20}.
Thus, instead of simple gradient averaging, we perform ``weighted aggregation'' in \texttt{OmniLearn} where we assign more weights to gradients from workers with larger batch-sizes.
We do so by \emph{scaling} the gradients on worker $k$ with batch-size $b_{k}$ relative to the global batch-size by a factor of $\lambda_{k}$, as shown in Equation~(\ref{eqn:bspweightedgrads}).
This preserves the global batch-size as well as convergence properties of distributed SGD, shown as Equation~(\ref{eqn:bspvariableSGD}) \cite{b21}.

\begin{table}[t]
	\centering
	\caption{BSP variable-batching convergence}
	\begin{tabular}{|c|c|c|c|}
		\hline
		\bfseries Model & \bfseries Iterations & \bfseries HL & \bfseries Test Accuracy \\
		\hline
		\multirow{3}{*}{ResNet18} & \multirow{3}{*}{$28 \times 10^{3}$} & \textit{HL1} & 95\% \\
		\cline{3-4}
		& & \textit{HL4} & 94.54\% \\
		\cline{3-4}
		& & \textit{HL8} & 94.03\% \\
		\hline
		\multirow{3}{*}{ResNet50} & \multirow{3}{*}{$76 \times 10^{3}$} & \textit{HL1} & 88\% \\
		\cline{3-4}
		& & \textit{HL4} & 86.6\% \\
		\cline{3-4}
		& & \textit{HL8} & 84.65\% \\
		\hline
		\multirow{3}{*}{AlexNet} & \multirow{3}{*}{$15 \times 10^{3}$} & \textit{HL1} & 70\% \\
		\cline{3-4}
		& & \textit{HL4} & 68.6\% \\
		\cline{3-4}
		& & \textit{HL8} & 68.15\% \\
		\hline
		\multirow{3}{*}{VGG11} & \multirow{3}{*}{$15 \times 10^{3}$} & \textit{HL1} & 83\% \\
		\cline{3-4}
		& & \textit{HL4} & 81.5\% \\
		\cline{3-4}
		& & \textit{HL8} & 79.04\% \\
		\hline
      \end{tabular}
	\label{table:bspVarBatchAccTTA}
\end{table}

\begin{figure}
	\hspace{-0.17cm}
	\begin{subfigure}{0.22\textwidth}
		\includegraphics[width=1.1\linewidth]{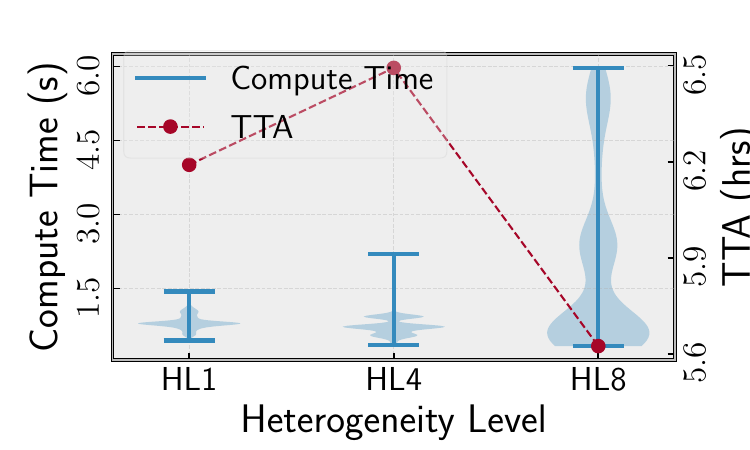}
   	 	\caption{ResNet18}
   	 	\label{fig:aspbaselineResnet18itrdensity}
	\end{subfigure}
	\hspace{1.5em}
	\begin{subfigure}{0.22\textwidth}
		\centering
		\includegraphics[width=1.1\linewidth]{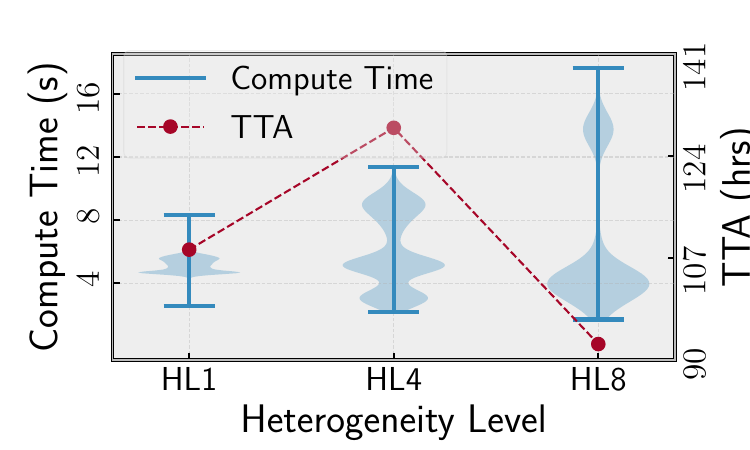}
		\caption{ResNet50}
		\label{fig:aspbaselineResnet50itrdensity}
  	\end{subfigure}
  	
  	\vspace{0.25em}
  	
  	\hspace{-0.17cm}
  	\begin{subfigure}{0.22\textwidth}
		\includegraphics[width=1.1\linewidth]{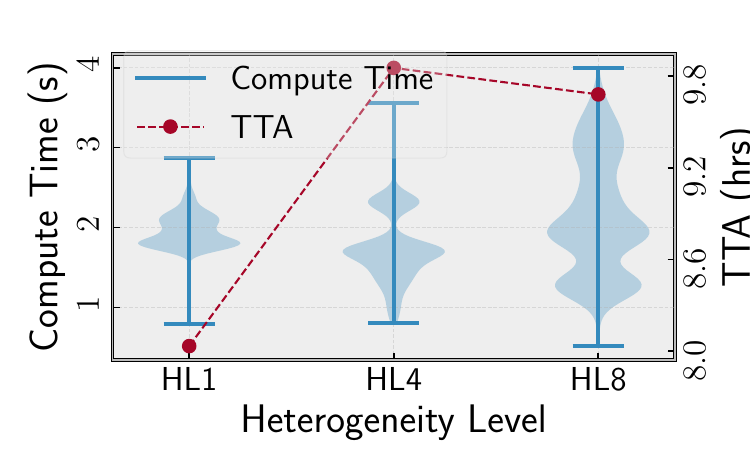}
   	 	\caption{AlexNet}
   	 	\label{fig:aspbaselineAlexnetitrdensity}
	\end{subfigure}
	\hspace{1.5em}
	\begin{subfigure}{0.22\textwidth}
		\centering
		\includegraphics[width=1.1\linewidth]{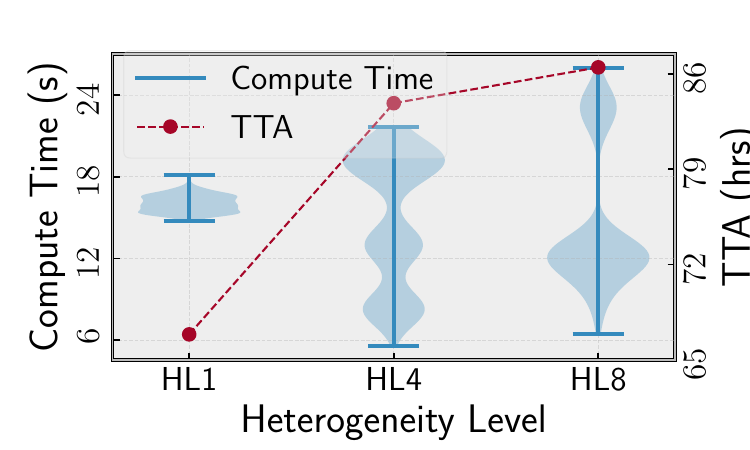}
		\caption{VGG11}
		\label{fig:aspbaselineVGGitrdensity}
  	\end{subfigure}
	\caption{The median compute time in BSP is lower with variable-batching, but the min-max is more spread apart at high \textit{HL}s as allocating batches based on core-count is not highly accurate for throughput estimation.}
	\label{fig:bspVarBatchingResults}
\end{figure}

BSP convergence quality with variable-batching at various \textit{HL}s is reported in Table~(\ref{table:bspVarBatchAccTTA}).
Please note that variable-batching at \textit{HL1} is equivalent to uniform-batching as workers have the same mini-batch size.
The general trend is that test accuracy decreases as heterogeneity increases.
As heterogeneity rises from \textit{HL1} to \textit{HL8}, workers with fewer resources are allocated smaller batches while faster nodes are assigned larger batches.
Computing updates from slow workers with smaller batches produces noisy gradients which can greatly degrade a model's statistical efficiency (i.e., test accuracy/loss) \cite{b19, b16}.
However, our weighted aggregation approach mitigates this by giving less importance to inaccurate gradients, and more value to updates with high signal-to-noise ratio calculated from larger batches.
Gradient scaling is thoroughly evaluated with variable-batching on heterogeneous clusters in \S \ref{eval:scaledunscaledgrads}.

As for parallel efficiency, Figure~(\ref{fig:bspVarBatchingResults}) shows the per-iteration compute time and time-to-accuracy (TTA) to reach the aforementioned targets.
For all models, the second quartile of compute-time distribution is lower at \textit{HL8} than \textit{HL1}.
This is because variable-batching attempts to reduce average computation time across the cluster by adjusting worker batches proportional to resource availability.
We make two interesting observations here: \emph{first, min-max compute-time over the distribution is more spread apart at higher \textit{HL}s.
Second, time to convergence increases for some models (AlexNet and VGG11) as we raise heterogeneity from \textit{HL1} to \textit{HL8}, and decreases for others (ResNet18 and ResNet50)
This is because allocating batches based simply on core-count is not a highly accurate throughput estimator and still results in some variance between worker compute times.
Additionally, batch execution is not strictly parallel over multi-core devices.
This means that allocating a higher batch-size to a worker with more cores does not necessarily mean its corresponding computation time will be proportionally smaller.}

For comparing statistical performance between uniform vs. variable-batching in BSP, accuracy in the former corresponds to \textit{HL1} in Table~(\ref{table:bspVarBatchAccTTA}), while the latter corresponds to each \textit{HL} entry in the same table.
As for parallel performance, worker compute time distributions and change in TTA over the \textit{HL}s can be compared between the two from Figures~(\ref{fig:bsdptimedistbaseline}) and (\ref{fig:bspVarBatchingResults}).
As \textit{HL} rises, TTA always increases in uniform-batching, while it may not necessarily be true in variable-batching.

\vspace{0.4em}

\subsubsection{Asynchronous Training}

For ASP, uniform-batching updates global model parameters by computing updates over $b$ samples at every iteration.
In static heterogeneity, we vary per-worker batches like Equation~(\ref{eqn:varbszBSP}) to reduce the effects of staleness in ASP training.
We do not perform gradient averaging in ASP as worker updates are not averaged concurrently over the PS.
As batch-size can greatly affect model generalization \cite{b22, b23}, we keep this parameter consistent between ASP and BSP.
The cumulative batch-size $\sum_{i=1}^{K}b_{k}$ is split among $K$ workers in accordance with Equation~(\ref{eqn:varbszBSP}) (even though updates are calculated from fewer samples $b_{k}$ of worker $k$ at step $i$ in Equation~(\ref{eqn:aspweightedgrads})).

To account for ASP's smaller mini-batches that produce noisier gradients \cite{b19, b16}, we also apply a linear learning-rate (LR) scaling rule, shown in Equation~(\ref{eqn:aspvariableSGD}), and use the LR $\eta_{k}$ on worker $k$ \cite{b24, b25}.
This is in harmony with prior work that explores trades-offs between tuning learning-rate or batch-size \cite{b34}.
In \S \ref{eval:LRASPVarBatchEval}, we compare the statistical performance of ASP under variable-batching \emph{with} and \emph{without} LR scaling.


\begin{subequations}
  \begin{equation}
    g_{(k, i)} = \dfrac{\partial}{\partial \theta_{i}}(\dfrac{1}{|b_{k}|} \sum_{b_{(i, k)} \in \mathcal{B}_{k}} \mathcal{F}(b_{(i, k)}, \theta_{i}))
    \label{eqn:aspweightedgrads}
  \end{equation}
  \begin{equation}
    \theta_{i+1}^{(k)} = \theta_{i}^{(k)} - \eta_{k} \odot g_{(k, i)} \;\Rightarrow\; \eta_{k} = \eta \cdot \dfrac{b_{k}}{\sum_{i=1}^{K}b_{i}}
    \label{eqn:aspvariableSGD}
  \end{equation}
  \label{eqn:ASPVarBatch}
\end{subequations}

\begin{table}[t]
	\centering
	\caption{ASP variable-batching performance}
	\begin{tabular}{|c|c|c|c|c|}
		\hline
		\bfseries Model & \bfseries Iterations & \bfseries HL & \bfseries Accuracy & \bfseries TTA (hrs)\\
		\hline
		\multirow{3}{*}{ResNet18} & \multirow{3}{*}{$120 \times 10^{3}$} & \textit{HL1} & 89.16\% & 22.2 \\
		\cline{3-5}
		& & \textit{HL4} & 90.18\% & 14.4 \\
		\cline{3-5}
		& & \textit{HL8} & 93.4\% & 9.6 \\
		\hline
		\multirow{3}{*}{ResNet50} & \multirow{3}{*}{$300 \times 10^{3}$} & \textit{HL1} & 82.64\% & 150.4 \\
		\cline{3-5}
		& & \textit{HL4} & 82.9\% & 168.4 \\
		\cline{3-5}
		& & \textit{HL8} & 61.4\% & 121.5 \\
		\hline
		\multirow{3}{*}{AlexNet} & \multirow{3}{*}{$30 \times 10^{3}$} & \textit{HL1} & 65.48\% & 8.1 \\
		\cline{3-5}
		& & \textit{HL4} & 65.77\% & 7.83 \\
		\cline{3-5}
		& & \textit{HL8} & 66.62\% & 7.81 \\
		\hline
		\multirow{3}{*}{VGG11} & \multirow{3}{*}{$48 \times 10^{3}$} & \textit{HL1} & 82\% & 107.3 \\
		\cline{3-5}
		& & \textit{HL4} & 81.53\% & 104.5 \\
		\cline{3-5}
		& & \textit{HL8} & 74.82\% & 104.6 \\
		\hline
      \end{tabular}
	\label{table:ASPVarBatchAccTTAs}
\end{table}

\begin{figure}
  \hspace{-0.2cm}
  \begin{subfigure}{0.22\textwidth}
    \includegraphics[width=1.1\linewidth]{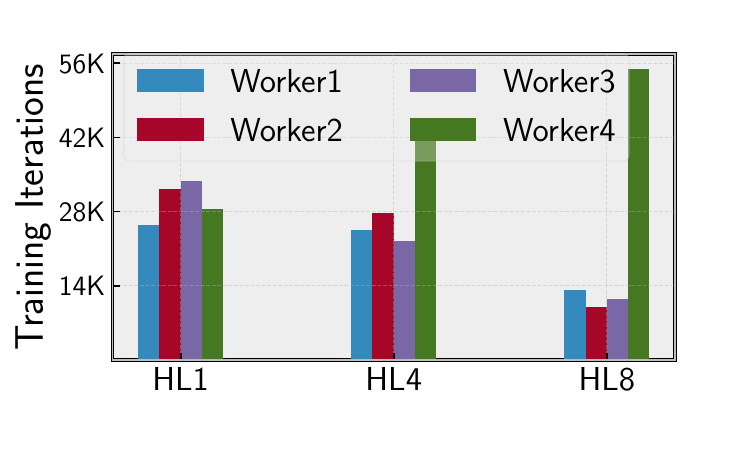}
    \caption{ResNet18}
    \label{fig:aspVarBatItrDensityRes18}
  \end{subfigure}
  \hspace{1.5em}
  \begin{subfigure}{0.22\textwidth}
    \centering
    \includegraphics[width=1.1\linewidth]{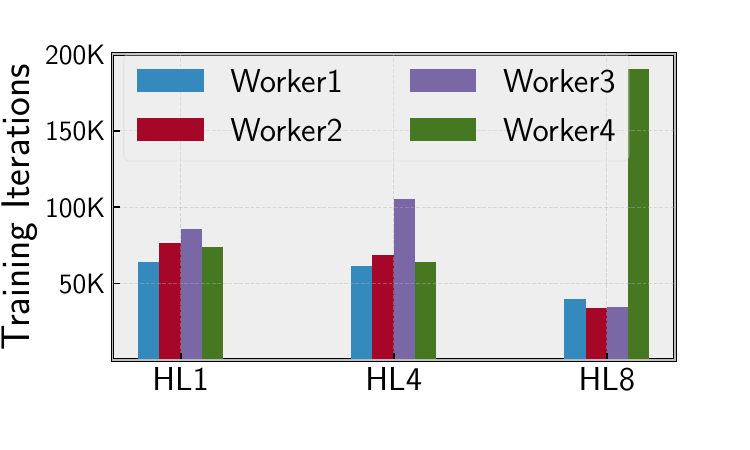}
    \caption{ResNet50}
    \label{fig:aspVarBatItrDensityRes50}
  \end{subfigure}
  
  \vspace{0.25em}
  
  \hspace{-0.2cm}
  \begin{subfigure}{0.22\textwidth}
    \includegraphics[width=1.1\linewidth]{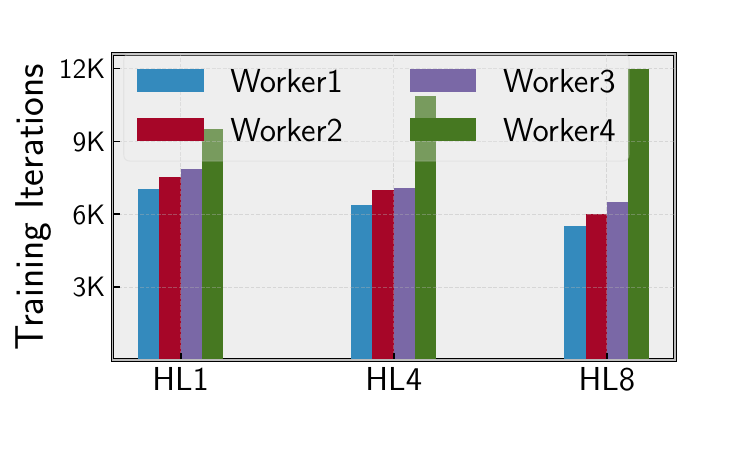}
    \caption{AlexNet}
    \label{fig:aspVarBatItrDensityAlex}
  \end{subfigure}
  \hspace{1.5em}
  \begin{subfigure}{0.22\textwidth}
    \centering
    \includegraphics[width=1.1\linewidth]{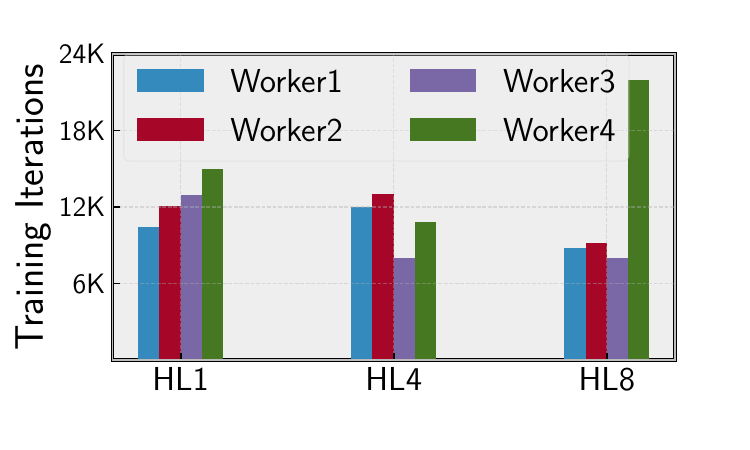}
    \caption{VGG11}
    \label{fig:aspVarBatItrDensityVGG}
  \end{subfigure}
  \caption{ASP worker density of model updates across \textit{HL}s.
    The accuracy degradation at high \textit{HL}s is attributed to staleness from skewed distribution of training iterations.}
  \label{fig:aspVariableBatchKDEs}
\end{figure}

We report the parallel and statistical performance of \texttt{OmniLearn}'s variable-batching technique under ASP training in Table~(\ref{table:ASPVarBatchAccTTAs}).
As before, we use varying degrees of heterogeneity by changing resource configurations in the 4-node cluster from Table~(\ref{table:computecoreHLs}).
In many cases, accuracy attained by the model improves even though heterogeneity gets worse, even surpassing ASP's homogeneous configuration.
For e.g.,  \textit{HL4} and \textit{HL8} in ResNet18 and AlexNet get better accuracy than \textit{HL1}, and so does \textit{HL4} in ResNet50.
This is because tweaking batches according to each node's capacity minimizes staleness, while asynchronous training allows for better exploration of worker's local minimas that improves model generalizability \cite{b10, b38}.
However, as heterogeneity gets even worse (to \textit{HL8}), models' statistical performance can suffer as well.
The accuracy of \textit{HL8} for ResNet50 and VGG11 is lower compared to \textit{HL1} or \textit{HL4}.
This is because staleness is not entirely eliminated even with variable-batching in certain training configurations due to throughput estimation errors.
However, compared to uniform batching (Figure~(\ref{fig:aspbaselineplotsitrDense})), there is considerable improvement in staleness in terms of iteration density distribution between fast and slow workers, as seen from Figure~(\ref{fig:aspVariableBatchKDEs}).

With the approximated work allocation policy of variable-batching, training time also decreases as we raise heterogeneity from \textit{HL1} to \textit{HL4} and \textit{HL8} as workers are able to process their respective local batches in lesser time.
The accuracy targets attained by each configuration is attributed to the respective staleness' in model updates, which is further corroborated in Figure~(\ref{fig:aspVariableBatchKDEs}) from workers' frequency distribution plots.
In highly heterogeneous scenarios like \textit{HL8}, ResNet50 and VGG11 lose considerable accuracy due to staleness from uneven worker updates (where worker \#4 performs most parameter updates in Figures~(\ref{fig:aspVarBatItrDensityRes50}) and (\ref{fig:aspVarBatItrDensityVGG}) since it has the most resources).
For the same models, \textit{HL4} has similar density distribution as \textit{HL1} which reflects in their convergence targets as well.
ResNet18 and AlexNet have better accuracy at higher \textit{HL}s as staleness is effectively mitigated in this case (as seen from similar iteration frequencies of workers), thus improving its statistical efficiency.

Uniform and variable-batching in ASP can be compared from Tables~(\ref{table:aspbaselineAccTTA}) and (\ref{table:ASPVarBatchAccTTAs}).
Compared to former, the rise in TTA with increasing \textit{HL}s is much less steeper in the latter and even decreases in some cases like ResNet18.

\emph{\textbf{Summary:} Both synchronous and asynchronous mechanisms benefit from splitting work as per our variable-batching approach.
Additional optimizations like weighted gradient aggregation in BSP and learning-rate scaling in ASP can further improve convergence quality in heterogeneous settings.
Compared to uniform-batching in \S \ref{subsec:designHeteroImpact}, BSP variable-batching improves parallel efficiency by attenuating straggler slowdown.
But in spite of gradient scaling, BSP's convergence quality is lower than uniform-batching due to noisier updates coming from small-batch nodes.
On the other hand, ASP's statistical efficiency is greatly improved with variable-batching across all heterogeneous configurations.
(compared to uniform-batching in Table~(\ref{table:aspbaselineAccTTA}).
However, the time to convergence may be slightly higher even at the same \textit{HL} as nodes may have higher compute times (from processing small batches on smaller workers and contrariwise).
This is because variable-batching does not always predict worker throughput accurately and batch execution over different neural networks is not strictly parallel over many-core processors.
}

It should also be noted that like uniform-batching, variable-batching with BSP has superior statistical performance than ASP.
This is due to the inherent convergent properties of synchronous SGD.
In some cases, even the time-to-accuracy in synchronous training may be better if the number of iterations taken to converge in ASP are excessively high.
We leave it to the ML practitioners to assess the trade-offs between BSP or ASP training based on their target objectives.
In this work, we present \texttt{OmniLearn} to work with either communication mechanism and accelerate training in heterogeneous clusters.

\subsection{Proportional Control based Dynamic-Batching}\label{subsec:propctrldynhet}

To mitigate stragglers or staleness, workers need to process their mini-batches simultaneously.
In the previously described static approach, mini-batches were allocated either based on server cores or FLOPs count.
However, this open-loop estimation of worker throughput is not accurate to predict the actual training throughput given a model and training hardware.
This is because throughput in distributed training depends on characteristics governed by parallel scaling laws.
Further, many scenarios yield dynamic resource availability for which the static approach is ill-suited.
To cope with this, the `controller' component in Figure~(\ref{fig:omnidesignfig}) implements Equation~(\ref{eqn:dynpropctrl}) for proportional control-based \textit{dynamic-batching}.


\begin{subequations}
  \begin{equation}
    T_{k}^{(e)} = \dfrac{b_{k}^{(e)}}{t_{k}^{(e)}} \:\; \text{and} \:\; \tau_{k}^{(e)} = t_{k}^{(e)} - \bar{t}^{(e)} :\: \bar{t}^{(e)} = \dfrac{1}{K} \sum_{k=1}^{K}t_{k}^{(e)}
    \label{eqn:itrtimeerror}
  \end{equation}
  \begin{equation}
    b_{k}^{(e+1)} = b_{k}^{(e)} + \Delta(b_{k}^{(e)}) \;\;:\;\; \Delta(b_{k}^{(e)}) = -\; T_{k}^{(e)}\tau_{k}^{(e)}
    \label{eqn:dynbszadjust}
  \end{equation}
  \label{eqn:dynpropctrl}
\end{subequations}	

\texttt{OmniLearn}'s dynamic-batching is designed to overcome throughput estimation errors, as well as dynamic resource availability due to overcommitment or performance interference from concurrent jobs on a server that results in fluctuating throughput (i.e., dynamic heterogeneity).
By frequently adjusting worker batch-size proportionally to a workers' throughput, we ensure that all workers incur roughly the same computation cost.
For worker `$k$' calculating its update in time $t_{k}^{(e)}$ and training throughput $T_{k}^{(e)}$ at epoch `$e$', we want $t_{i}^{(e)} = t_{j}^{(e)} \; \forall \; (i, j)$ workers.
Given the average iteration time across all workers as $\bar{t}^{(e)}$, dynamic adjustment uses a simplistic proportional-control approach to equalize batch processing times, i.e., minimize the \emph{error} $\tau_{k}^{(e)}$ shown in Equation~(\ref{eqn:itrtimeerror}).
This error is minimized by updating worker batch-size according to Equation~(\ref{eqn:dynbszadjust}).
Slower workers thus have a positive error $\tau_{k}^{(e)}$ and need to decrease their effective batch-size while faster workers have a negative $\tau_{k}^{(e)}$ and increase their local batch-size since they are capable of handling a higher workload.
Our approach essentially combines model-based and conventional black-box PID controllers.
Instead of tuning an arbitrary constant as used commonly, we use the estimated throughput.
Dynamic-batching also works with any initial batch-size.
By default, workers can be assigned their respective batch-size based on static heterogeneity and any errors in throughput approximations (based on CPU/GPU cores, FLOPs etc.) are corrected by the control mechanism.

\vspace{0.4em}
\subsubsection{Control Stability for Dynamic-Batching}

We can perform dynamic-batching as part of \texttt{OmniLearn} at any granularity, say every few steps or epochs.
However, we tune worker batches at the end of every epoch in our evaluation since batch-adjustment is not a zero-cost operation as it involves checkpointing, re-batching and re-initializing the PyTorch dataloader or TensorFlow estimator with updated batches.
Also, the computation time on workers will rarely converge to the exact average, and thus there will always be some degree of error that the proportional control mechanism will try to chase.
To alleviate this perpetual batch-size adjustment and data re-partitioning with frequently updated batches, we leverage three main techniques: \textit{dead-banding}, \textit{exponentially smoothed computation times} and \textit{batch-size bounds}.

\vspace{0.2em}
\textbf{\emph{Dead-Banding}}: Currently, \texttt{OmniLearn} calculates the updated batches at the end of every epoch via proportional-control.
On top of that, we integrate deadbanding with our controller; batch-sizes are updated only when the change is substantial and surpasses a specified threshold ($\Delta b_{k}$).
With Equation~(\ref{eqn:omnibszdeadbandlimit}), batch-size on worker $k$ is updated to $b_{k}^{(e+1)}$ if:

\begin{equation}
  \Delta b_{k} \leq \bigg|\dfrac{b_{k}^{(e+1)} - \; b_{k}^{(e)}}{b_{k}^{(e+1)}}\bigg|
  \label{eqn:omnibszdeadbandlimit}
\end{equation}

When the change in batch-size is lower than $\Delta b_{k}$, no batch re-adjustments are made.
The threshold is chosen based on how sensitive we want the adjustments to be, in addition to the performance overhead of adapting the batch-size.

\vspace{0.2em}
\textbf{\emph{Exponential smoothing}}: With dead-banding, \texttt{OmniLearn} makes batch adjustments when the underlying resource availability changes.
To improve controller stability and avoid spurious adjustments, we compute the error $\tau_{k}$ (i.e., deviation between worker computation time and the average cluster computation time).
We then apply EWMA smoothing (\textbf{E}xponentially \textbf{W}eighted \textbf{M}oving \textbf{A}verage) where higher priority is assigned to later iterations.
This provides us with an integrator component in the controller which prevents outliers.

\vspace{0.2em}
\textbf{\emph{Batch-size bounds}}: As already discussed, batch-size is an important hyperparameter in deep learning.
With dynamic-batching, the slowest worker may have a very small batch-size, while batches may become colossal on a fast worker in extremely heterogeneous systems and degrade test accuracy \cite{b22, b23}.
Additionally, proportional control may even reduce the overall training throughput due to constraint-free varying of batches in dynamic-batching.
BSP throughput may drop either from training with smaller batches (due to poor parallelization), or from large batch-sizes significantly increasing a worker's computation time.
Furthermore, very small or large batches are not optimal from a parallel efficiency standpoint given an execution hardware.
Small batches cannot leverage maximal parallelism, while large batches may exhaust memory resources and result either in low throughput or job termination.
To mitigate this, we enforce lower-upper batch-size bounds ($b_{min}, b_{max}$) in heterogeneous training.

\begin{equation}
	b_{k}^{(e+1)} = b_{k}^{(e+1)} - \dfrac{\triangle B}{K} \;\;:\;\; \triangle B = \sum_{k=1}^{K}b_{k}^{(e+1)} - Kb
	\label{eqn:dynbszGlobalBFixed}
\end{equation}

Despite min-max batch-size constraints, the global batch-size can rise significantly over subsequent adjustments in BSP training, especially when heterogeneity varies dynamically and frequently.
\emph{We thus enforce an additional rule in BSP to keep the global batch-size fixed to reduce generalization loss and fairly compare with uniform-batching (where global batch-size is $Kb$ for $K$ workers each with mini-batch $b$)}.
We do so by proportionately reducing adjusted batches (from Equation~(\ref{eqn:dynpropctrl})) relative to deviation from global batch $Kb$ (in Equation~(\ref{eqn:dynbszGlobalBFixed})).

\begin{figure}
  \hspace{-1em}
  \begin{subfigure}{0.22\textwidth}
    \includegraphics[width=1.1\linewidth]{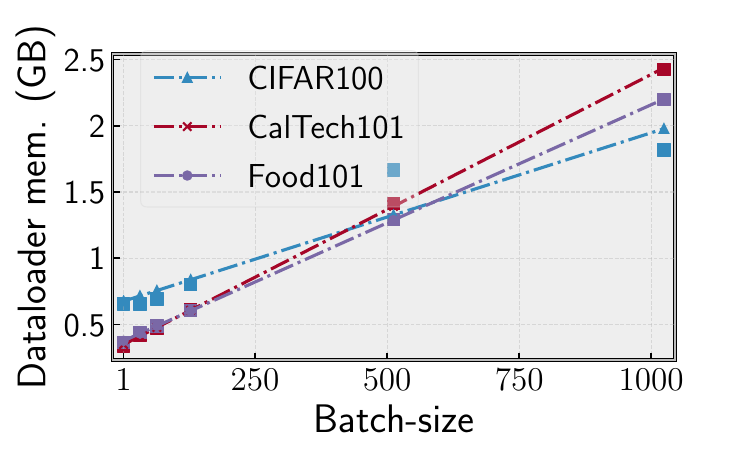}
    \caption{\scriptsize{$M_{batch}$ predicted vs. actual}}
    \label{fig:dataloaderMem}
  \end{subfigure}
  \hspace{1em}
  \begin{subfigure}{0.22\textwidth}
    \centering
    \includegraphics[width=1.1\linewidth]{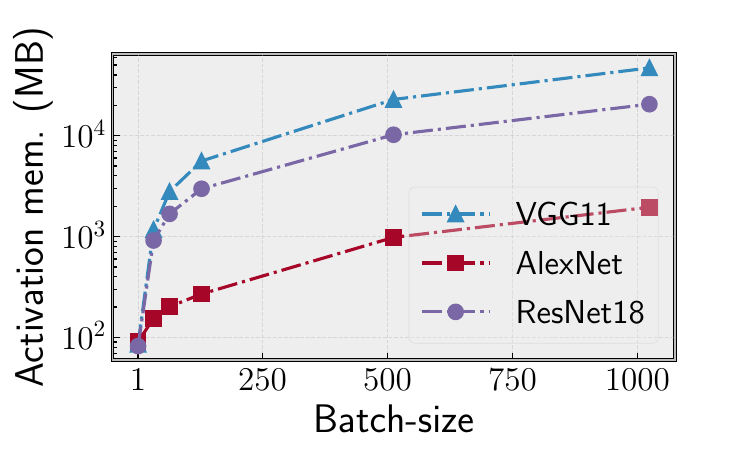}
    \caption{\scriptsize{Activation memory}}
    \label{fig:activationMem}
  \end{subfigure}
  \caption{(a) $M_{batch}$ for any batch-size is accurately predicted by the linear model. (b) $M_{act} \propto$ batch-size (y-axis is log-scale).}
  \label{fig:memUtils}
\end{figure}

\vspace{0.4em}
\subsubsection{Heterogeneous memory across workers}\label{subsubsec:hetmemOmnilearn}

Rather than the size or quantity of compute units, heterogeneity may also arise from different types and amounts of memory across workers.
The total memory needed for training involves storing model-parameters, gradient, activation and batch-memory \cite{b35}, as shown in Equation~(\ref{eqn:hetmemeqn}).

\begin{equation}
	M_{total} = M_{model} + M_{grad} + M_{opt} + M_{act} + M_{batch}
	\label{eqn:hetmemeqn}
\end{equation}

$M_{model}$ and $M_{grad}$ is fixed for a given model and calculated as the product of total trainable parameters and byte-size per parameter (e.g., 4 bytes for single-precision, 2 bytes for half-precision, etc.).
$M_{opt}$ is the optimizer memory that varies with the type and information stored by the optimizer, yet proportional to the model-size and remains fixed throughout training.
For e.g., memory held by a plain SGD optimizer is significantly lower than that needed by Nesterov's momentum and Adam optimizer \cite{b35}.
The last two, $M_{batch}$ and $M_{act}$ vary with batch-size and thus, may limit the maximum batch-size we can choose.
The latter is the memory held by activation functions computed on each batch sample in forward pass, while the former is the memory consumed by the dataloader to process/store samples corresponding to a chosen batch-size.

We fit the relationship between $M_{batch}$ and batch-size as a linear regression problem in Figure~(\ref{fig:dataloaderMem}).
For CIFAR100, CalTech101 and Food101 datasets, the dashed line shows the estimated memory by our model while the scatter points show the actual memory consumed solely by dataloaders at batches of 1, 32, 64, 128, 512 and 1024.
Activation memory is also a linear function of batch-size that can be readily estimated from a model's structure and batch-size.
To validate the same, we run VGG11, AlexNet and ResNet18 on the $M_{batch}$ datasets mentioned earlier with a plain SGD optimizer.
We deduct $M_{model}$, $M_{grad}$, $M_{opt}$ and $M_{batch}$ from the total memory consumed by the training process.
The first three are the same as the model-size itself, while we fetched $M_{batch}$ for a specific batch-size from Fig. (\ref{fig:dataloaderMem}).
For a given batch-size, this leaves only the activation memory which we plot in Figure~(\ref{fig:activationMem}).
The y-axis is on logscale and shows $M_{act}$ linearly increasing with the batch-size.
With this, we can accurately predict the memory required to run a specific model configuration, and accurately set the cap for min-max batch-size to use over nodes with heterogeneous memory in \texttt{OmniLearn}.

\vspace{0.4em}
\subsubsection{Putting it all together}

\begin{algorithm}
  \DontPrintSemicolon
  \SetKwProg{Pn}{procedure}{:}{\KwRet}
  
  \SetKwFunction{train}{Train}
  \SetKwFunction{range}{range}
  \SetKwFunction{centralized}{centralized}
  \SetKwFunction{decentralized}{decentralized}
  \SetKwFunction{time}{time}
  \SetKwFunction{ewma}{EWMA}
  \SetKwFunction{reduce}{AllReduce}
  \SetKwFunction{summ}{SUM}
  \SetKwFunction{sendPS}{sync.pushToPS}
  \SetKwFunction{recvPS}{sync.pullFromPS}
  \SetKwFunction{batchdata}{batchTrainingData}
  \SetKwFunction{dynBatch}{getBsz}
  \SetKwFunction{avgcomputetime}{$\bar{t}^{compute}_{(i, k)}$}
  \SetKwFunction{collectitrrtime}{getAvgComputeTime}
  \caption{Synchronous training in \texttt{OmniLearn}}
  \textbf{Input:} $K$, $B$, $E$, $\eta$, $\theta_{(0)}$, $\Delta b_{k}$, $b_{min}$, $b_{max}$ \\
  \Pn{\train{}}{
    $b_{(1, k)} = B / K$ \\
    \For {$\text{epoch} \; e \; \text{in}$ \range{$1, E$} \text{on worker k}}{
      $\lambda_{k} = b_{(e, k)} / B$ \\
      $\mathcal{B} = $ \batchdata{$b_{(e, k)}$} \\
      \For {$\text{iteration} \; i \: \text{in} \; 0,1,2,... \;$}{
        $t^{(init)} = \time{}$ \\
        $g_{(i, k)} = \dfrac{\partial}{\partial \theta_{i}}(\dfrac{1}{|b_{(e, k)}|} \sum_{\mathit{\mathcal{B}_{(i, k)}}} \mathcal{F}(\mathcal{B}_{(i, k)},\theta_{i}))$ \\
        $\bar{t}^{(compute)}_{(i, k)}$ = \ewma{$(\time{} - t^{(init)})$} \\
        $\bar{g}_{(i)} = \reduce{$\lambda_{k} \odot g_{(i, k)}$, \summ}$ \\
        \BlankLine
        $\theta_{(i+1)} = \theta_{(i)} - \eta \cdot \bar{g}_{(i)}$ \\
      }
      \BlankLine
      $\bar{t}^{(avg)}_{(e)} = $ \collectitrrtime{\avgcomputetime} \\
      \vspace{0.05cm}
      $b_{(e+1, k)}$ = \dynBatch{$b_{(e, k)}, \bar{t}^{(compute)}_{(i, k)}, \bar{t}^{(avg)}_{(e)}$}
    }
  }
  \vspace{0.08cm}
  \BlankLine
  \Pn{\dynBatch{$b_{(e, k)}, \bar{t}^{(compute)}_{(i, k)}, \bar{t}^{(avg)}_{(e)}$}}{
    $T_{(e, k)} = b_{(e, k)} \: / \: \bar{t}^{(compute)}_{(i, k)}$ \\
    \vspace{0.07cm}
    $b_{(e+1, k)}^{(*)} = b_{(e, k)} - T_{(e, k)}(\bar{t}_{(i, k)}^{(compute)} - \bar{t}^{(avg)}_{(e)})$ \\
    \If{$\Big|\dfrac{b_{(e+1, k)}^{(*)} - b_{(e, k)}}{b_{(e+1, k)}^{(*)}}\Big| \geq \Delta b_{k}$}{
      \vspace{0.12cm}
      \If{$b_{(e+1, k)}^{(*)} < b_{min}$}{
        $b_{(e+1, k)} = b_{min}$ \\
      } \ElseIf{$b_{(e+1, k)}^{(*)} > b_{max}$}{
        $b_{(e+1, k)} = b_{max}$ \\
      } \Else{
        $b_{(e+1, k)} = b_{(e+1, k)}^{(*)}$ \\
      }
    }
    \textbf{return} $b_{(e+1, k)}$
  }
  \label{algo:omniBSPdynamicrun}
\end{algorithm}

Algorithm~(1) shows how \texttt{OmniLearn} integrates control stability with proportional control in BSP.
All workers initially use a uniform batch-size of $B/K$.
The function \texttt{batchTrainingData} generates a random mini-batch of the specfiied size from the training dataset.
For controller stability, we use exponential moving average of the gradient computation time (line 10). 
This is followed by gradient aggregation step via \texttt{AllReduce} collective in a decentralized topology.

Each workers' gradients are scaled by $\lambda_{k}$ for weighted averaging in BSP.
In our implementation, dynamic-batching is applied at the end of each epoch, although this granularity can easily be changed.
Procedure \texttt{getAvgComputeTime} is another communication step to collect and average the computation times across all workers. 
The overhead of this step is low as only a single 32-bit floating point value is communicated by each worker. 
The procedure \texttt{getBsz} evaluates each worker's potential batch-size as per Equation~(\ref{eqn:dynpropctrl}) and tweaks it only if threshold $\Delta b_{k}$ is exceeded.
Batches are chosen accordingly based on min-max batch-size bounds, while keeping $B$ fixed. 

\begin{algorithm}
  \DontPrintSemicolon
  \SetKwProg{Pn}{procedure}{:}{\KwRet}
  \SetKwProg{with}{with}{:}{\KwRet}
  \SetKwFunction{train}{Train}
  \SetKwFunction{range}{range}
  \SetKwFunction{centralized}{centralized}
  \SetKwFunction{time}{time}
  \SetKwFunction{ewma}{EWMA}
  \SetKwFunction{aa}{$\alpha$}
  \SetKwFunction{summ}{SUM}
  \SetKwFunction{pushPS}{async.pushToPS}
  \SetKwFunction{pullPS}{async.pullFromPS}
  \SetKwFunction{batchdata}{batchTrainingData}
  \SetKwFunction{workerTimes}{wrkTime}
  \SetKwFunction{workerxx}{wrkTime[$k$]}
  \SetKwFunction{workerxy}{wrkTime[$wid$]}
  \SetKwFunction{workerBsz}{wrkB}
  \SetKwFunction{workbszxx}{wrkB[$k$]}
  \SetKwFunction{workbszxy}{wrkB[$wid$]}
  \SetKwFunction{processbsz}{evalBsz}
  \SetKwFunction{size}{sizeOf}
  \SetKwFunction{mean}{mean}
  \SetKwFunction{avgcomputetime}{$\bar{t}^{compute}_{(e, k)}$}
  \SetKwFunction{collectitrrtime}{getAvgComputeTime}
  \SetKwFunction{dynBatch}{getBsz}
  \SetKwFunction{onworker}{\textbf{On Workers:}}
  \SetKwFunction{onps}{\textbf{On Parameter Server:}}
  \caption{Asynchronous training in \texttt{OmniLearn}}
  \textbf{Input:} $K$, $B$, $E$, $\eta$, $\theta_{(0)}$, $\Delta b_{k}$, $b_{min}$, $b_{max}$ \\
  \vspace{0.05cm}	
  \textbf{On Workers:} \\
  \Pn{\train{}}{
    $b_{(1, k)} = B / K$ \\
    \For {$\text{epoch} \; e \; \text{in}$ \range{$1, E$} \text{on worker k}}{
      $\eta_{k} = \eta \cdot b_{(e, k)} \:/\: B$ \\
      $\mathcal{B} = $ \batchdata{$b_{(e, k)}$} \\
      \For {$\text{iteration} \; i \: \text{in} \; 0,1,2,... \;$}{
        $\theta_{(i)} =$ \pullPS{} \\
        $t^{(init)} = \time{}$ \\
        $g_{(i, k)} = \dfrac{\partial}{\partial \theta_{i}}(\dfrac{1}{|b_{(e, k)}|} \sum_{\mathit{\mathcal{B}_{(i, k)}}} \mathcal{F}(\mathcal{B}_{(i, k)},\theta_{i}))$ \\
        $\bar{t}^{(compute)}_{(e, k)}$ = \ewma{$(\time{} - t^{(init)})$} \\
        $\theta_{(i+1, k)} = \theta_{(i)} - \eta_{k} \cdot g_{(i, k)}$ \\
        \vspace{0.05cm}
        \pushPS{$\theta_{(i+1, k)}$} \\
      }
      \BlankLine
      $b_{(e+1, k)}$ = \processbsz{$k, b_{(e, k)}, \bar{t}^{(compute)}_{(e, k)}$}
    }
  }
  \hrulefill \\
  \vspace{0.1cm}
  \textbf{On Parameter Server:}	 \\
  \Pn{\pullPS{}}{
    \textbf{return} $\theta_{(i)}$ \\
  }
  \BlankLine
  \Pn{\pushPS{$\theta_{(i+1, k)}$}}{
    $\theta_{(i+1)} = \theta_{(i+1, k)}$ \\
  }
  \BlankLine
  \Pn{\processbsz{$k, b_{(e, k)}, \bar{t}^{(compute)}_{(e, k)}$}}{
    \workbszxx = $b_{(e, k)}$ \\
    \vspace{0.02cm}
    \workerxx = $\bar{t}^{(compute)}_{(e, k)}$ \\
    \If{\size{\workerBsz} = $K$}{
      $\bar{t}^{(avg)}_{(e)}$ = \mean{\workerTimes .values$()$} \\
      \For{$wid$, $b$ in \workerBsz .items$()$}{
        $\bar{t}^{(compute)}_{(e, k)}$ = \workerxy \\
        $b_{(e+1, k)}$ = \dynBatch{$b$, $\bar{t}^{(compute)}_{(e, k)}$, $\bar{t}^{(avg)}_{(e)}$}
        \workbszxy = $b_{(e+1, k)}$
      }
      \vspace{0.04cm}
      \textbf{return} \workbszxx
    }
  }
  \label{algo:omniASPdynamicrun}
\end{algorithm}

\texttt{OmniLearn}'s pseudocode for ASP training is described in Algorithm~(2). 
Unlike BSP that scales gradients proportional to worker batch-size, we don't apply gradient scaling in ASP since only a single worker updates model parameters on the central PS in a given iteration.
Instead, we scale the learning-rate based on a worker's batch-size (line 6).
The latest model parameters are pulled from the PS at the beginning of every iteration (via procedure \texttt{pullFromPS}) in order to compute local updates.
The computation time is logged (line 12), followed by applying SGD update on local model replica (line 13) and sending updated model back to the PS.
The synchronization cost of ASP is low as there is no straggler slowdown since workers push their updates asynchronously.
A worker makes remote call to the PS via function \texttt{evalBsz} and transmits its worker-id $k$, local batch-size and computation time.
The PS holds the latter two in a hashmap/dictionary with a key corresponding to every worker (lines 23-24), computes average cluster computation time (line 26) and applies control stability similar to \texttt{getBsz} for BSP in Algorithm~(1). 
\emph{The average computation time for a worker measured over an epoch may vary slightly between BSP and ASP, as the number of samples accumulated also varies between the two.}
In BSP, compute-time is logged at each iteration for every worker, while high-resource workers in ASP perform more iterations on the global PS model and have more samples to calculate the average computation time compared to low-resource workers.


\section{Experimental Evaluation}\label{sec:eval}

\subsection{Implementation and Cluster Setup}\label{subsec:clustermodeldescribe}

We evaluate \texttt{OmniLearn} over PyTorch v2.2.0, with centralized ASP implemented over PyTorch RPC, and decentralized BSP via MPI using DDP module \cite{bb78}.
We also validate over TensorFlow Estimator v1.15.1 with a centralized implementation.
In our experiments, we measure the parallel and statistical performance of our approach in settings with static and dynamic heterogeneity.
We compare with vanilla PyTorch distributed (i.e., uniform-batching) in ASP and BSP training, and report final model accuracy, worker compute times and TTA improvements.
The following training schedules are employed to run various models under uniform-batching as well as \texttt{OmniLearn}'s variable and dynamic-batching.

We use a variety of image classification datasets to evaluate \texttt{OmniLearn}, ASP and BSP training.
Food101 \cite{b27} consists of 101 food categories with 101,000 images, with 250 test images and 750 training images available per-class.
CalTech101 \cite{b29} comprises of 9000 object images from 101 classes, with about 40-800 images available per-class.
The size of each image is roughly 300$\times$200 pixels.
CalTech256 \cite{b30} contains 256 object categories across 30,607 images.
Over CalTech101, this dataset has more than 2$\times$ classes, contains at least 80 images per-label, and avoids artifacts from image rotation.
Places365-Standard dataset \cite{b32} consists of about 1.8 million images from 365 scene categories, with up to 5000 images per-category.
Models hyperparameters are:

\begin{itemize}
	\item ResNet18 \cite{b26} is trained on Food101 dataset with worker batch-size 32, SGD with momentum 0.9 and weight decay 5e-4, initial learning-rate (LR) 0.1 that decays by a factor of 10 after 15, 30 and 45 epochs respectively.
	\item AlexNet \cite{b28} trains with batch-size 32 on CalTech101 using momentum 0.9, weight decay 5e-4, and initial LR 0.01 decaying by 0.1 at epochs 80, 120 and 160.
	\item ResNet50 \cite{b26} trains on CalTech256 with worker batch-size 32, momentum 0.9, weight decay 1e-4, initial LR 0.1, and decay factor 0.1 at 100, 150 and 200-\textit{th} epoch.
	\item VGG11 \cite{b31} runs on Places365-Standard with batch-size 32, momentum 0.9, weight decay 5e-4, initial LR 0.01 with decay-rate 0.2 at epochs 15, 25 and 35 respectively.
\end{itemize}

We evaluate \texttt{OmniLearn} over Docker v23.0.5 running on a 48-core Intel Xeon E5-2670 @2.3GHz Ubuntu v20.04.6 system with 384 GB memory.
Spawning docker containers as separate workers of a cluster ensures minimal interference, resource isolation and emulates various heterogeneity scenarios crucial for our evaluation.
We simulate different \textit{HL} configurations with respect to compute cores by mapping and limiting a container to a specific set of CPU-cores \cite{b33}, and allocate 32 GB memory to each container.
For centralized ASP, a PS container is allocated 8 CPU-cores with 32 GB memory.
To enable communication between workers, we launch docker swarm over a 10 Gbps network interconnect.

\subsection{Weighted aggregation in BSP variable-batching}\label{eval:scaledunscaledgrads}

\begin{table}
	\centering
	\caption{Weighted aggregation in BSP variable-batching}
	\begin{tabular}{|c|c|c|c|c|}
		\hline
		\multicolumn{2}{|c|}{\bfseries Configuration} &
      	\multicolumn{3}{c|}{\bfseries Test accuracy} \\
      	\hline
      	\bfseries Model & \bfseries HL & \bfseries Scaled grads. & \bfseries Unscaled grads. & \bfseries Diff. \\
      	\hline
      	\multirow{3}{*}{ResNet18} & \textit{HL2} & 94.9\% & 93.6\% & \textbf{+1.3\%} \\
      	\cline{2-5}
      	& \textit{HL4} & 94.54\% & 93.76\% & \textbf{+0.78\%} \\
      	\cline{2-5}
      	& \textit{HL8} & 94.02\% & 92.11\% & \textbf{+1.91\%} \\
      	\hline
      	\multirow{3}{*}{ResNet50} & \textit{HL2} & 87.74\% & 86\% & \textbf{+1.74\%} \\
      	\cline{2-5}
      	& \textit{HL4} & 86.6\% & 84.33\% & \textbf{+2.27\%} \\
      	\cline{2-5}
      	& \textit{HL8} & 84.65\% & 80.6\% & \textbf{+4.05\%}  \\
      	\hline
      	\multirow{3}{*}{AlexNet} & \textit{HL2} & 68.9\% & 65.1\% & \textbf{+4.8\%} \\
      	\cline{2-5}
      	& \textit{HL4} & 68.6\% & 60.67\% & \textbf{+7.93\%} \\
      	\cline{2-5}
      	& \textit{HL8} & 68.15\% & 57.94\% & \textbf{+10.21\%}  \\
      	\hline
      	\multirow{3}{*}{VGG11} & \textit{HL2} & 82.38\% & 79.4\% & \textbf{+2.98\%} \\
      	\cline{2-5}
      	& \textit{HL4} & 81.5\% & 75.6\% & \textbf{+5.9\%} \\
      	\cline{2-5}
      	& \textit{HL8} & 79.04\% & 71.19\% & \textbf{+7.85\%} \\
      	\hline
      \end{tabular}
	\label{table:bspscaledunscaledgrads}
\end{table}

For synchronous training, \texttt{OmniLearn} performs \emph{weighted aggregation} or \emph{gradient scaling} where the contribution of each node is weighted according to the batch-size processed by it (instead of a simply averaging updates).
Thus, more importance is given to updates coming from large-batch nodes, as gradients calculated from smaller batches tend to be noisier while those from larger-batches tend to be more aligned with the true gradients \cite{b19, b16}.
To see the efficacy of this approach, we compare BSP variable-batching \textit{with} and \textit{without} gradient scaling in Table~(\ref{table:bspscaledunscaledgrads}); the latter computes a simple mean by summing up the gradients and dividing by the total cluster-size.
\emph{The results show that gradient scaling considerably improves \texttt{OmniLearn}'s convergence quality over unscaled gradients; by up to 1.91\%, 4.05\%, 10.21\% and 7.85\% in ResNet18, ResNet50, AlexNet and VGG11 respectively}.
This is shown in the `Diff.' column that gives the delta between the accuracy with scaled and unscaled gradients, which gets even larger at high heterogeneity settings like \textit{HL4} and \textit{HL8}. 
Our weighted aggregation assigns higher priority to less noisy gradients, thus improving BSP's statistical performance.

\begin{table}
	\centering
	\caption{Learning-rate scaling in ASP variable-batching}
	\begin{tabular}{|c|c|c|c|c|}
		\hline
		\multicolumn{2}{|c|}{\bfseries Configuration} &
      	\multicolumn{3}{c|}{\bfseries Test accuracy} \\
      	\hline
      	\bfseries Model & \bfseries HL & \bfseries w/ LR scaling & \bfseries w/o LR scaling & \bfseries Diff. \\
      	\hline
      	\multirow{2}{*}{ResNet18} & \textit{HL4} & 90.18\% & 88.62\% & \textbf{+1.56\%} \\
      	\cline{2-5}
      	& \textit{HL8} & 93.4\% & 91.35\% & \textbf{+2.05\%} \\
      	\hline
      	\multirow{2}{*}{ResNet50} & \textit{HL4} & 82.9\% & 80.64\% & \textbf{+2.26\%} \\
      	\cline{2-5}
      	& \textit{HL8} & 61.4\% & 54.5\% & \textbf{+6.9\%} \\
      	\hline
      	\multirow{2}{*}{AlexNet} & \textit{HL4} & 65.77\% & 63.61\% & \textbf{+2.16\%} \\
      	\cline{2-5}
      	& \textit{HL8} & 66.62\% & 65.36\% & \textbf{+1.26\%} \\
      	\hline
      	\multirow{2}{*}{VGG11} & \textit{HL4} & 81.53\% & 81.47\% & \textbf{+0.06\%} \\
      	\cline{2-5}
      	& \textit{HL8} & 74.82\% & 74.52\% & \textbf{+0.3\%} \\
      	\hline
      \end{tabular}
	\label{table:aspvarbatchLRscaling}
\end{table}

\subsection{Learning-rate scaling in ASP variable-batching}\label{eval:LRASPVarBatchEval}

Like batch-size, learning-rate (LR) is another crucial training hyperparameter.
Prior work has proposed LR tuning relative to cluster-size in data-parallel training \cite{b25, b24}.
For asynchronous training in \texttt{OmniLearn}, we adjust LR relative to the global batch-size had we trained synchronously (i.e., processed $Kb$ samples instead of $b$ per-step).
By doing so, we try to keep Equation~(\ref{eqn:ASPVarBatch}) independent of the effects of different batch-sizes, i.e., if batch $|b_{k}|$ in the denominator of Equation~(\ref{eqn:aspweightedgrads}) is changed by a certain fraction, then so is the learning-rate in Equation~(\ref{eqn:aspvariableSGD}). 
This approach follows the intuition that smaller batches prefer a smaller step-size to reach minima, and vice-versa \cite{b19}.
In Table~(\ref{table:aspvarbatchLRscaling}), we record the final accuracy at varying degrees of heterogeneity as we scale LR proportional to worker batch-size vs. use the default LR routine described in \S \ref{subsec:clustermodeldescribe}.
\emph{For ResNet18, ResNet50 and AlexNet, we observed considerable improvement in statistical performance by using a smaller step-size on workers training on smaller batches, and a relatively higher LR on large-batch workers (via LR scaling).
For an over-parameterized network like VGG11, we observed only marginal gains with LR scaling.
Specifically, ResNet18, ResNet50 and AlexNet attained up to 2.05\%, 6.9\% and 2.16\% improvement in accuracy, while the larger VGG11 achieved about 0.3\% accuracy gain}.

\subsection{Worker compute-cost distribution with different techniques}

\begin{figure}
	\hspace{-0.1cm}
	\begin{subfigure}{0.22\textwidth}
		\includegraphics[width=1.1\linewidth]{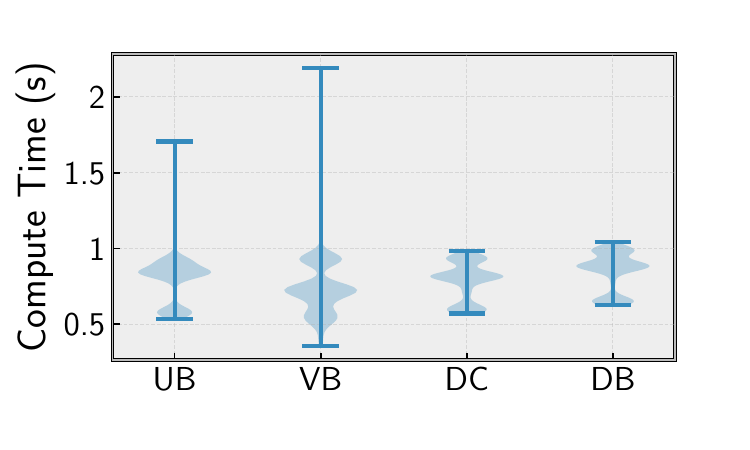}
   	 	\caption{ResNet18}
   	 	\label{fig:bspbaselineResNet18}
	\end{subfigure}
	\hspace{1.5em}
	\begin{subfigure}{0.22\textwidth}
		\centering
		\includegraphics[width=1.1\linewidth]{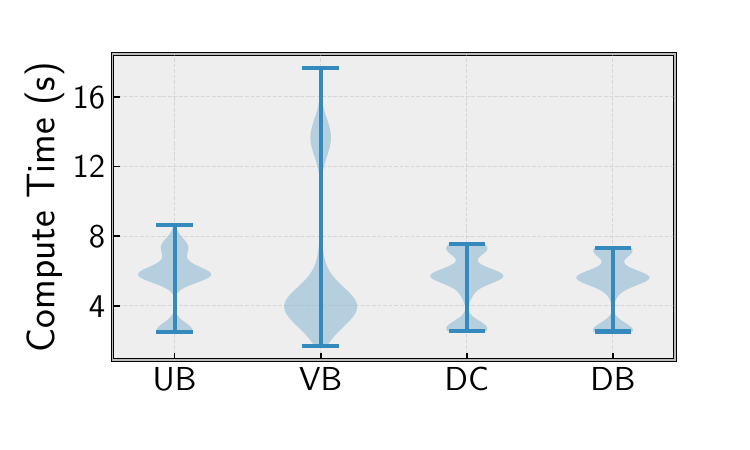}
		\caption{ResNet50}
		\label{fig:bspbaselineResNet50}
  	\end{subfigure}
  	
  	\vspace{0.25em}
  	\hspace{-0.1cm}
  	\begin{subfigure}{0.22\textwidth}
		\includegraphics[width=1.1\linewidth]{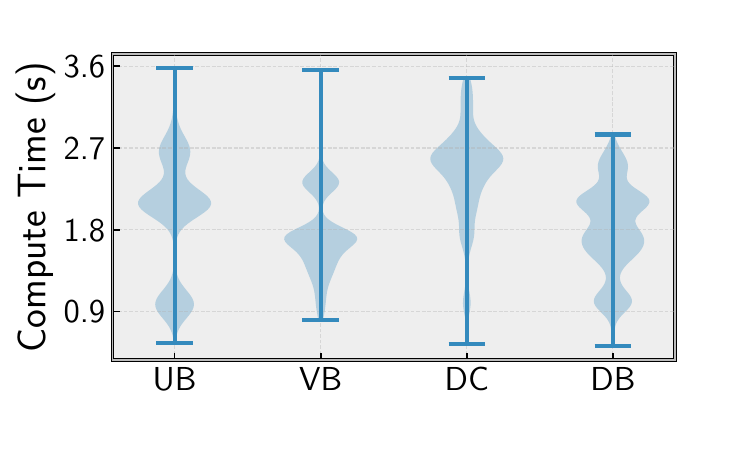}
   	 	\caption{AlexNet}
   	 	\label{fig:bspbaselineAlexNet}
	\end{subfigure}
	\hspace{1.5em}
	\begin{subfigure}{0.22\textwidth}
		\centering
		\includegraphics[width=1.1\linewidth]{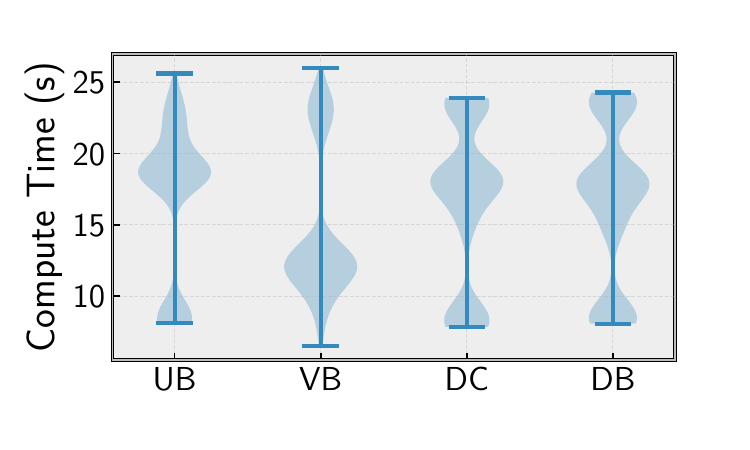}
		\caption{VGG11}
		\label{fig:bspbaselineVGG}
  	\end{subfigure}
	\caption{Distribution of worker computation times in Uniform-Batching (UB) and \texttt{OmniLearn}'s Variable-Batching (VB), Dynamic-Controller (DC) and DeadBand-controller (DB).}
	\label{fig:omnicomputecostdist}
\end{figure}

In this section, we compare worker compute-time distributions in BSP with uniform-batching (UB) and \texttt{OmniLearn}'s variable-batching (VB), dynamic-control (DC) and deadbanding (DB), shown in Figure~(\ref{fig:omnicomputecostdist}).
As we move from UB to VB, the quality of throughput estimation improves, and average cluster computation time is reduced.
Thus, the median or second quartile of compute-time density gets smaller as we move from uniform to variable-batching.
For e.g., ResNet18 with uniform-batching is most dense around 0.85 sec. and 0.64 sec. with variable-batching.
With ResNet50, UB is most dense at around 6s and VB at approximately 4s; AlexNet takes about 2.15s and 1.8s with UB and VB respectively.
The VGG11 (largest of the four) model observed maximum improvement.
Here, uniform-batching is clustered around 18.7s, while VB is centered at around 12.5s.
The maximum compute cost in VB can be as high as UB (in AlexNet and VGG11), or even higher (in ResNet18 and ResNet50).
This is due to poor throughput estimation by \emph{statically} allocating larger-batches to workers with more cores even though it does not fully parallelize.
With VB, we observed that the maximum compute-time was observed on workers with more compute resources (i.e., running larger batches).
We employ dynamic-control to mitigate this issue, which reduced the maximum compute-time observed and condensed the density distribution much better than VB.
With deadbanding, we apply the same proportional-control technique as DC, and further enforce a min-max constraint on the batch-size allocated to each worker.

\subsection{Deadband-controller under static heterogeneity}

\begin{figure*}
  \centering
  \begin{subfigure}{0.22\textwidth}
    \includegraphics[width=1.1\linewidth]{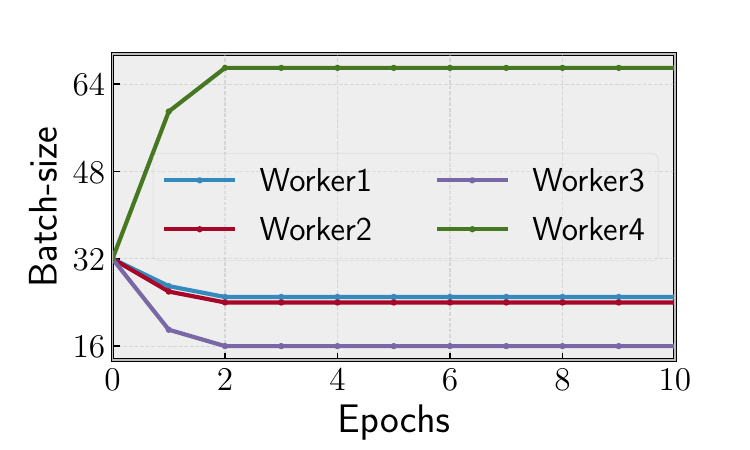}
    \caption{ResNet18 batch-sizes}
    \label{fig:dynbatAdjustBszRes18}
  \end{subfigure}
  \hfill
  \begin{subfigure}{0.22\textwidth}
    \centering
    \includegraphics[width=1.1\linewidth]{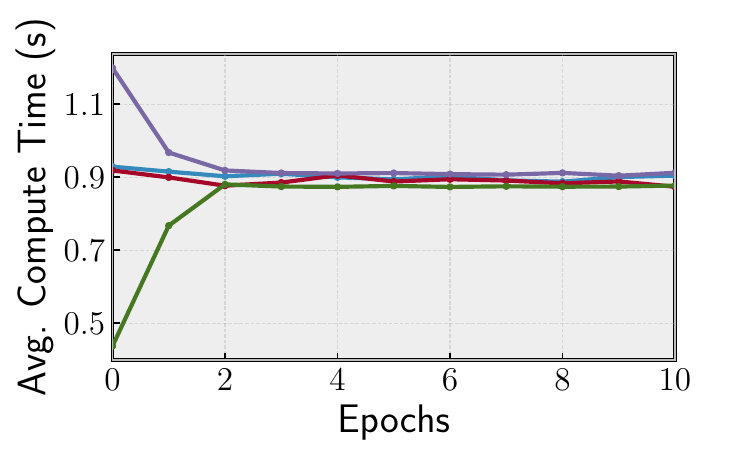}
    \caption{ResNet18 compute-times}
    \label{fig:dynbatAdjustTimeRes18}
  \end{subfigure}
  \hfill
  \begin{subfigure}{0.22\textwidth}
    \includegraphics[width=1.1\linewidth]{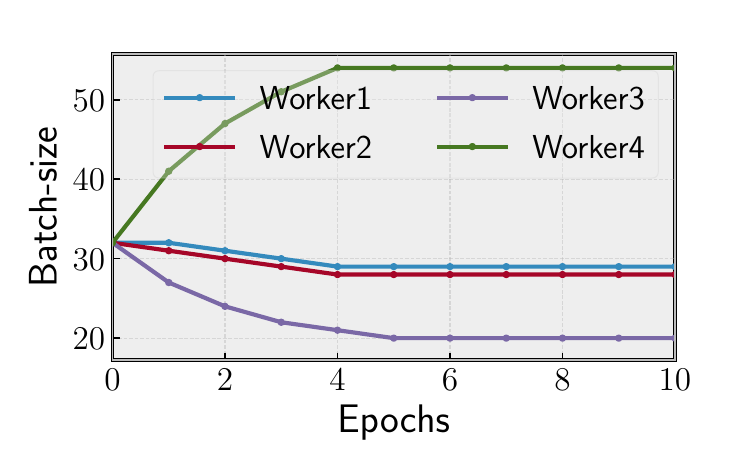}
    \caption{ResNet50 batch-sizes}
    \label{fig:dynbatAdjustBszRes50}
  \end{subfigure}
  \hfill
  \begin{subfigure}{0.22\textwidth}
    \includegraphics[width=1.1\linewidth]{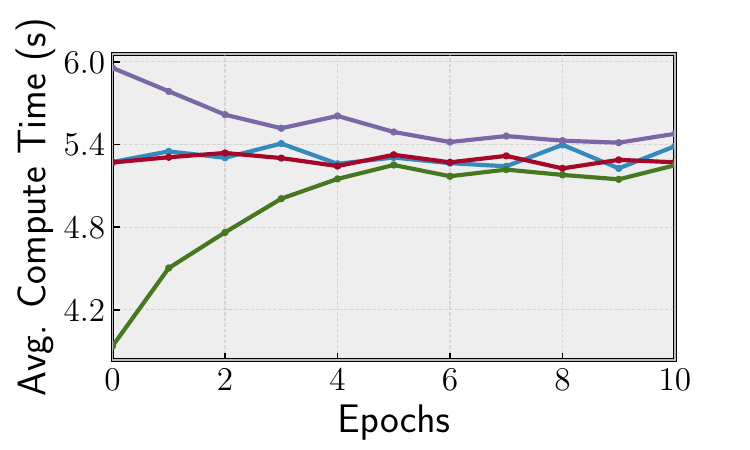}
    \caption{ResNet50 compute-times}
    \label{fig:dynbatAdjustTimeRes50}
  \end{subfigure}
  
  \vspace{1em}

  \begin{subfigure}{0.22\textwidth}
    \includegraphics[width=1.1\linewidth]{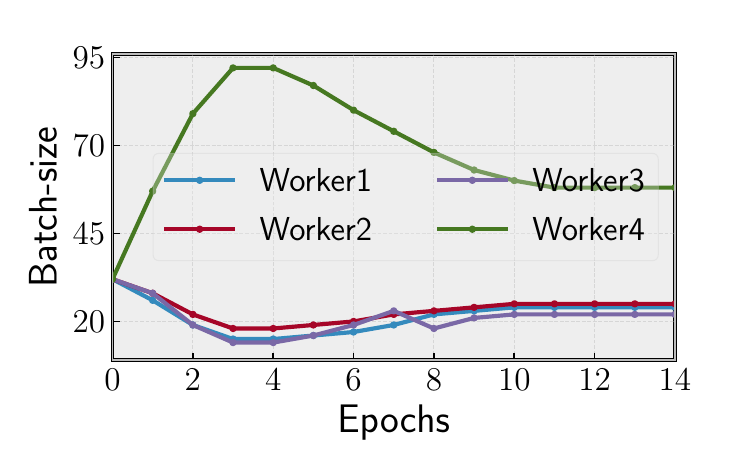}
    \caption{AlexNet batch-sizes}
    \label{fig:dynbatAdjustBszAlex}
  \end{subfigure}
  \hfill
  \begin{subfigure}{0.22\textwidth}
    \centering
    \includegraphics[width=1.1\linewidth]{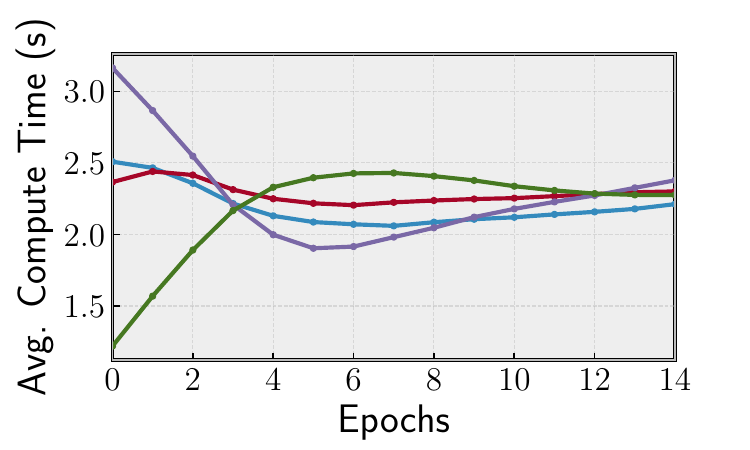}
    \caption{AlexNet compute-times}
    \label{fig:dynbatAdjustTimeAlex}
  \end{subfigure}
  \hfill
  \begin{subfigure}{0.22\textwidth}
    \includegraphics[width=1.1\linewidth]{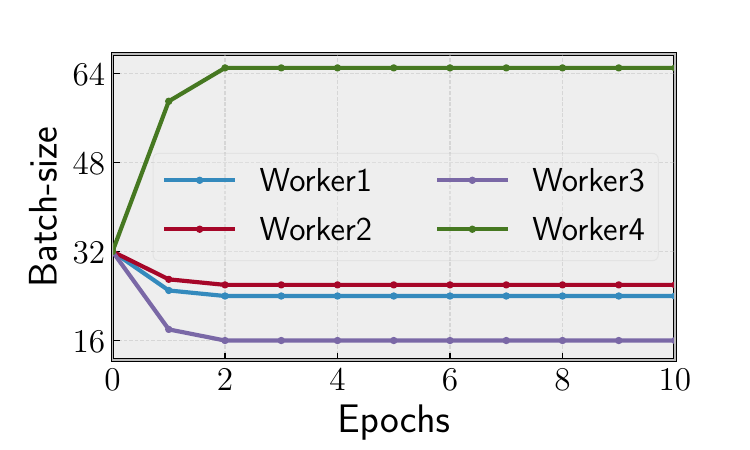}
    \caption{VGG11 batch-sizes}
    \label{fig:dynbatAdjustBszVGG}
  \end{subfigure}
  \hfill
  \begin{subfigure}{0.22\textwidth}
    \includegraphics[width=1.1\linewidth]{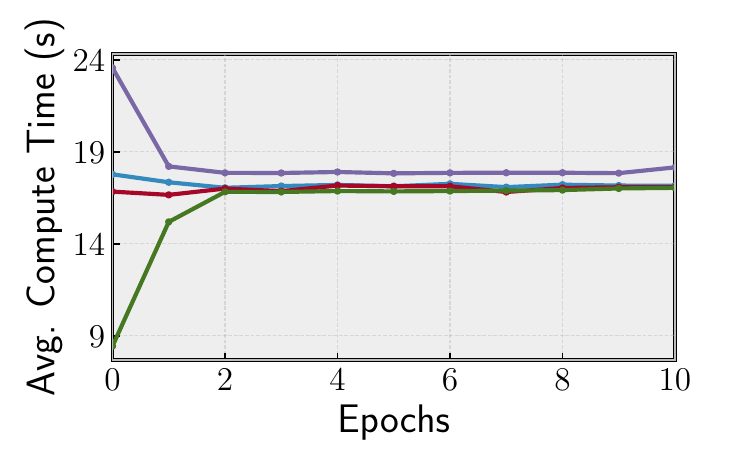}
    \caption{VGG11 compute-times}
    \label{fig:dynbatAdjustTimeVGG}
  \end{subfigure}
  
  \caption{\texttt{OmniLearn} BSP batches and compute-times under \emph{static} heterogeneity \emph{without} global batch-size constraints.}
  \label{fig:dynbatchBszAdjust}
\end{figure*}

We now see how well \texttt{OmniLearn}'s proportional-control works with any initial mini-batch under static heterogeneity.
Ideally, we want to allocate initial worker batches based on the open-loop variable-batching technique, and any throughput estimation errors are corrected over the course of training through its control mechanism.
Although a good initiation point is not mandatory for appropriate batch-size estimation in \texttt{OmniLearn}, the further the initial batch-size is from the ideal (i.e., proportional to worker throughput), the more adjustments are required to arrive at the steady-state equilibrium batches.
At \textit{HL8}, we perform BSP dynamic-batching with the same initial mini-batch 32 across all workers, and show batch adjustments and average computation times in the early epochs until the mini-batches stabilize.
Here, we set the min-max batches to (12, 96), $\triangle b_{k} = 15\%$, and perform proportional-control over BSP training (as explained in Equation~(\ref{eqn:dynpropctrl})) without constraints on the global batch-size (as imposed by Equation~(\ref{eqn:dynbszGlobalBFixed})).
Thus, the global batch-size may increase beyond 32$\times$4=128 if excessive batch adjustments are made in the 4-node cluster.
As we see in Figure~(\ref{fig:dynbatchBszAdjust}), \texttt{OmniLearn} need not make several batch-size tuning under static heterogeneity.
With proportional-control triggered at the end of each epoch, we see that ResNet18, ResNet50 and VGG11 easily converged to their respective steady-state batches in about 2-4 adjustments.
However, AlexNet took up to 10 adjustments before stabilizing as worker \#4 with most compute resources (allocated as per Table~(\ref{table:computecoreHLs})) peaked to its maximum allowable mini-batch governed by min-max bounds, then decreases and stabilizes to the nearest equilibrium state.
Despite the lack of global-batch constraints, \texttt{OmniLearn} increases the global mini-batch by only 12\% for ResNet50, and under 2.5\% for ResNet18, AlexNet and VGG11.
When heterogeneity varies dynamically and changes drastically with time, it may be necessary to impose global mini-batch bounds and keep the global batch-size fixed to preserve final model accuracy.

\emph{Compared to uniform-batching running a fixed batch-size under a static \textit{HL}, \texttt{OmniLearn} reduces TTA by up to 1.8 hrs in ResNet18, 8.6 hrs in ResNet50, 2.16 hrs in AlexNet and 20.3 hrs in VGG11}.

\begin{figure}[h]
  \hspace{-0.48cm}
  \begin{subfigure}{0.22\textwidth}
    \includegraphics[width=1.1\linewidth]{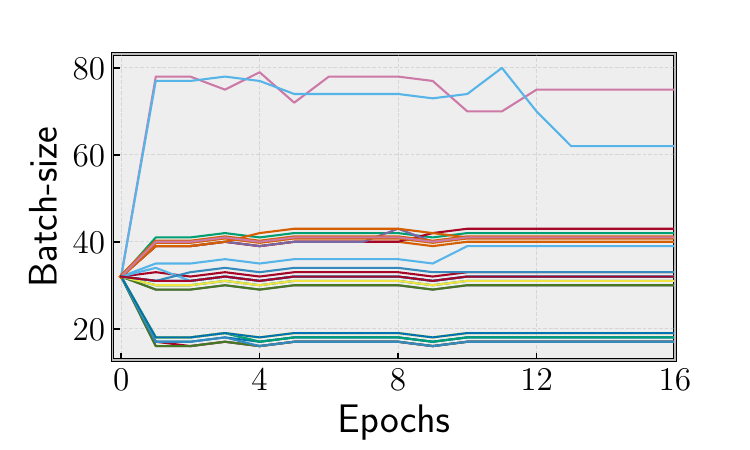}
    \caption{ResNet18 batch-sizes}
    \label{fig:bspdbandMinMaxRes1832VMbsz}
  \end{subfigure}
  \hspace{1.5em}
  \begin{subfigure}{0.22\textwidth}
    \centering
    \includegraphics[width=1.1\linewidth]{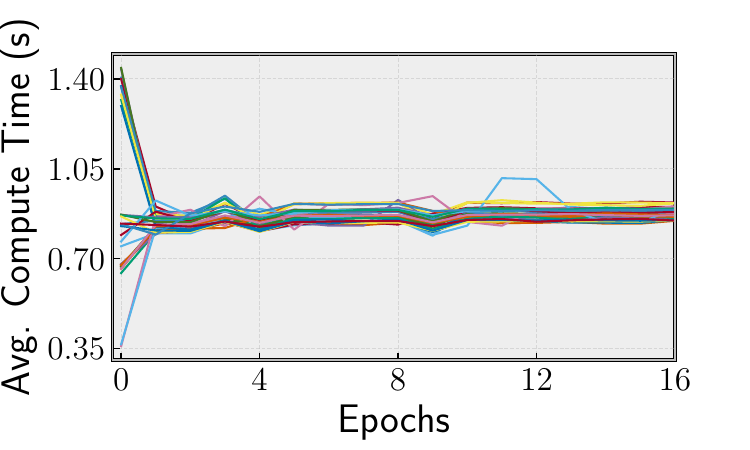}
    \caption{ResNet18 compute-times}
    \label{fig:bspdbandMinMaxRes1832VMitrtime}
  \end{subfigure}
  \hspace{-0.48cm}
  \begin{subfigure}{0.22\textwidth}
    \includegraphics[width=1.1\linewidth]{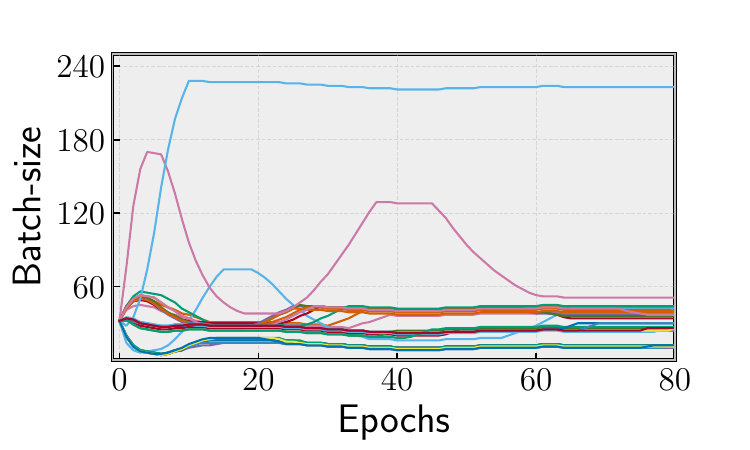}
    \caption{AlexNet batch-sizes}
    \label{fig:bspdbandMinMaxAlex32VMbsz}
  \end{subfigure}
  \hspace{1.5em}
  \begin{subfigure}{0.22\textwidth}
    \centering
    \includegraphics[width=1.1\linewidth]{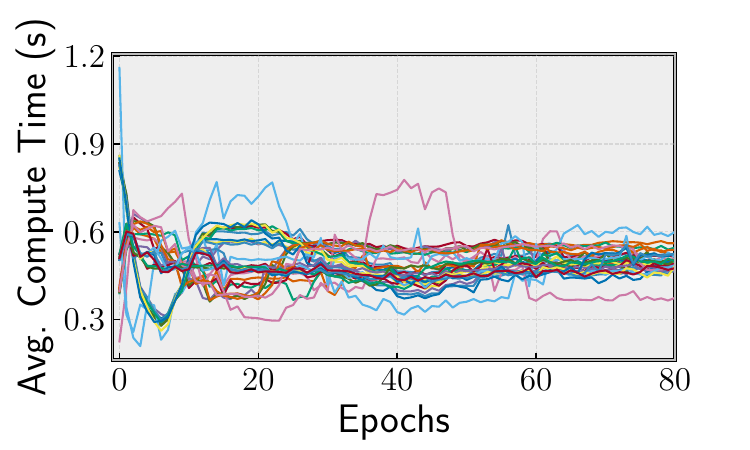}
    \caption{AlexNet compute-times}
    \label{fig:bspdbandMinMaxAlex32VMitrtime}
  \end{subfigure}
  \caption{\texttt{OmniLearn} training over a cluster of 32 workers.}
  \label{fig:bspdbandMinMax32VMs}
\end{figure}

\texttt{OmniLearn} is efficacious even over large, heterogeneous clusters.
The overhead of calculating batches is relatively low irrespective of cluster-size since only a floating-point (i.e., average compute time) is shared by each worker for proportional control.
We plot the batch-size adjustment and corresponding computation times for ResNet18 and AlexNet across 32 workers in Figure~(\ref{fig:bspdbandMinMax32VMs}).
To emulate heterogeneity across four servers each with 48-cores, a cluster of 32 containers are allocated such that 10 workers are composed of 2, 4 and 6 cores each, while the largest two workers are allocated 16 and 32-cores respectively.
From Figure~(\ref{fig:bspdbandMinMaxRes1832VMbsz}) and (\ref{fig:bspdbandMinMaxAlex32VMbsz}), the larger workers train over large batches while low-resource workers use small batches.
Similar to cluster-size 4, any change in a worker's average compute time triggers a corresponding change in its batch-size.
However, by the scaling the global batch-size proportional to cluster-size in BSP, 32 workers take 8$\times$ fewer steps to process the same cumulative samples as a cluster of 4 workers.
In that case, the training speedup comes from running fewer iterations for the same number of epochs.

\subsection{Deadband-controller under dynamic heterogeneity}

\begin{figure}
  \centering
  \begin{subfigure}{0.22\textwidth}
    \includegraphics[width=1.1\linewidth]{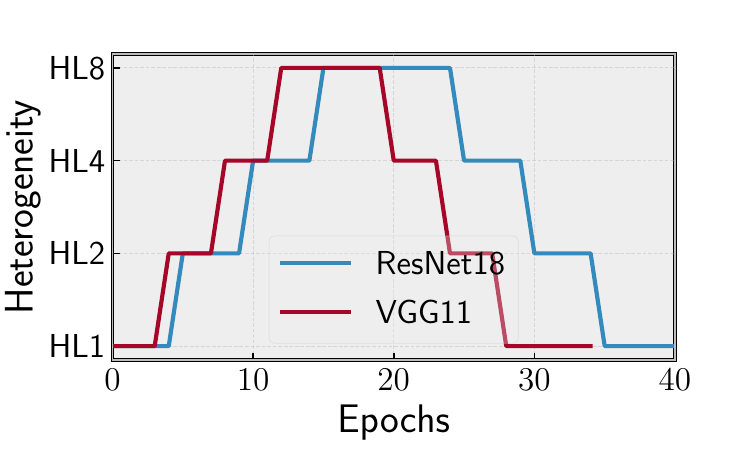}
    \caption{ResNet18 and VGG11}
    \label{fig:res18vggdynheterosim}
  \end{subfigure}
  \hspace{1.5em}
  \begin{subfigure}{0.22\textwidth}
    \centering
    \includegraphics[width=1.1\linewidth]{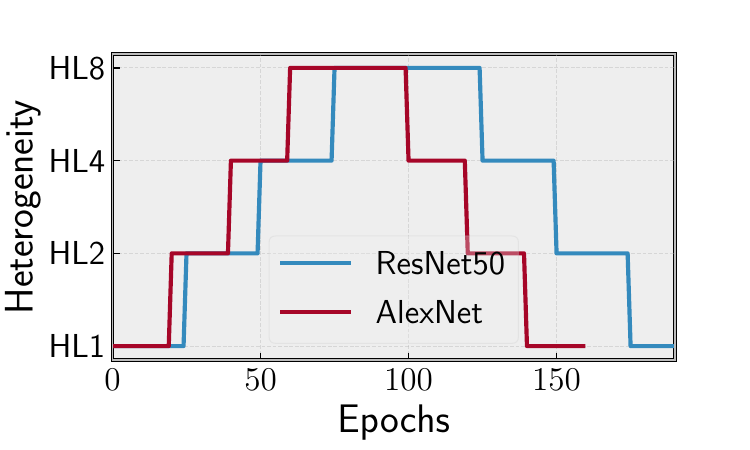}
    \caption{ResNet50 and AlexNet}
    \label{fig:res50alexdynheterosim}
  \end{subfigure}
  
  \caption{HLs are varied over the epochs to emulate dynamic heterogeneity in a cluster.}
  \label{fig:dynheterogensimulation}
\end{figure}

Dynamic-batching with control stability works well in static heterogeneity.
With frequent checks based on proportional-control, \texttt{OmniLearn} is also adept at responding to varying \textit{HL}s when computational resources fluctuate over time.
We simulate dynamic heterogeneity in a \emph{stepwise} manner between \textit{HL1}, \textit{HL2}, \textit{HL4} and \textit{HL8}.
From Figure~(\ref{fig:dynheterogensimulation}), we initiate \textit{HL} change at fixed intervals, triggered by an external process that tweaks the CPU-cores allocated to each worker container.
For a specific \textit{HL}, we allocate CPU core-sets for the 4 workers as per Table~(\ref{table:computecoreHLs}).
In ResNet18, we vary heterogeneity once every 5 epochs, and every 25 epochs in ResNet50.
With AlexNet and VGG11, we switch \textit{HL}s after every 20 and 4 epochs respectively.
At these granularities, we can scale heterogeneity from \textit{HL1} through \textit{HL8}, and back to \textit{HL1} over each models' entire training routine.
With this approach, we can emulate scenarios when resource availability fluctuates, akin to concurrent jobs deployed on a shared cluster, or mimic instances of burstable or spot VMs in the cloud.

\begin{figure*}[h]
  \centering
  \begin{subfigure}{0.22\textwidth}
    \includegraphics[width=1.1\linewidth]{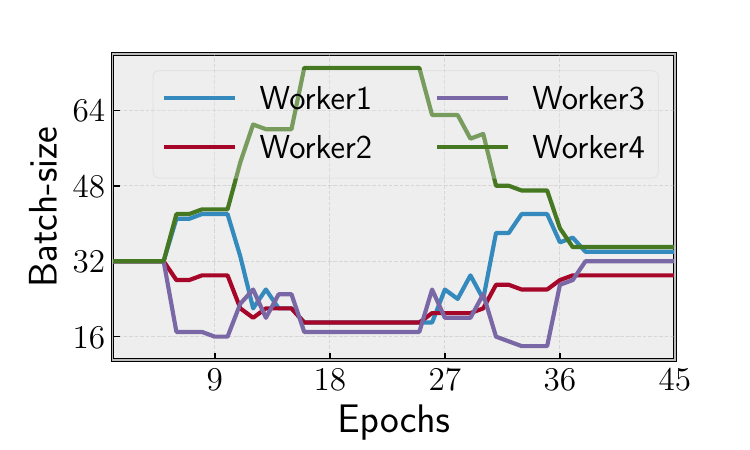}
    \caption{ResNet18 batch-sizes}
    \label{fig:dynbatBSPbszRes18}
  \end{subfigure}
  \hfill
  \begin{subfigure}{0.22\textwidth}
    \centering
    \includegraphics[width=1.1\linewidth]{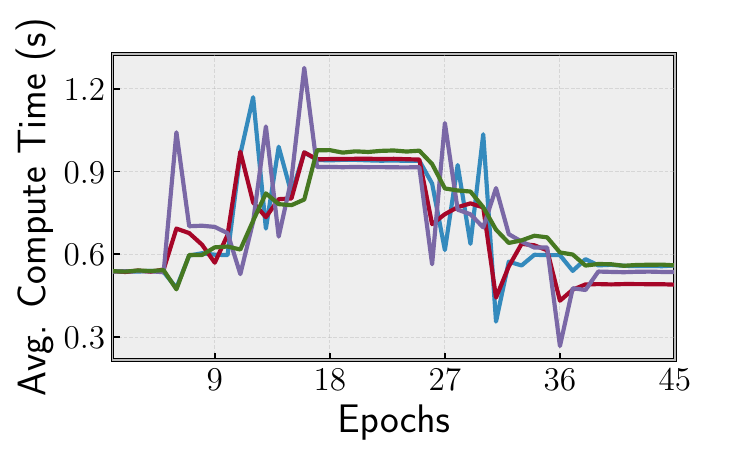}
    \caption{ResNet18 compute-times}
    \label{fig:dynbatBSPtimeRes18}
  \end{subfigure}
  \hfill
  \begin{subfigure}{0.22\textwidth}
    \includegraphics[width=1.1\linewidth]{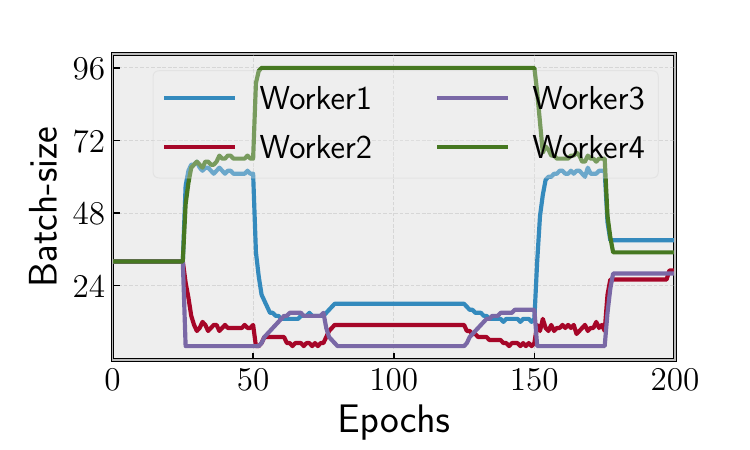}
    \caption{ResNet50 batch-sizes}
    \label{fig:dynbatBSPbszRes50}
  \end{subfigure}
  \hfill
  \begin{subfigure}{0.22\textwidth}
    \includegraphics[width=1.1\linewidth]{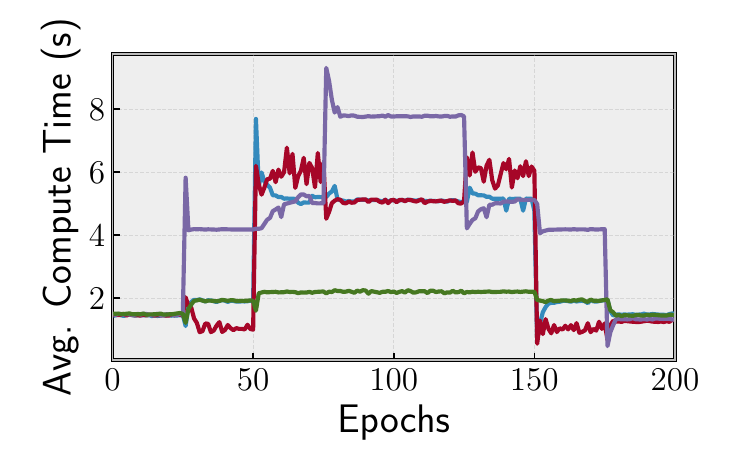}
    \caption{ResNet50 compute-times}
    \label{fig:dynbatBSPtimeRes50}
  \end{subfigure}
  
  \vspace{1em}

  \begin{subfigure}{0.22\textwidth}
    \includegraphics[width=1.1\linewidth]{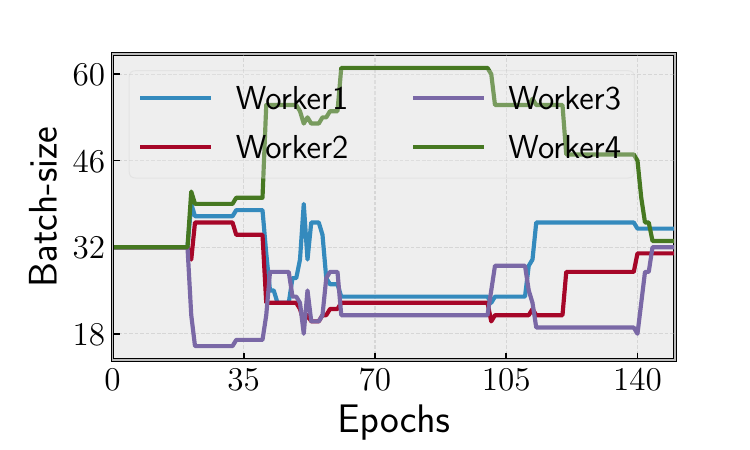}
    \caption{AlexNet batch-sizes}
    \label{fig:dynbatBSPbszAlex}
  \end{subfigure}
  \hfill
  \begin{subfigure}{0.22\textwidth}
    \centering
    \includegraphics[width=1.1\linewidth]{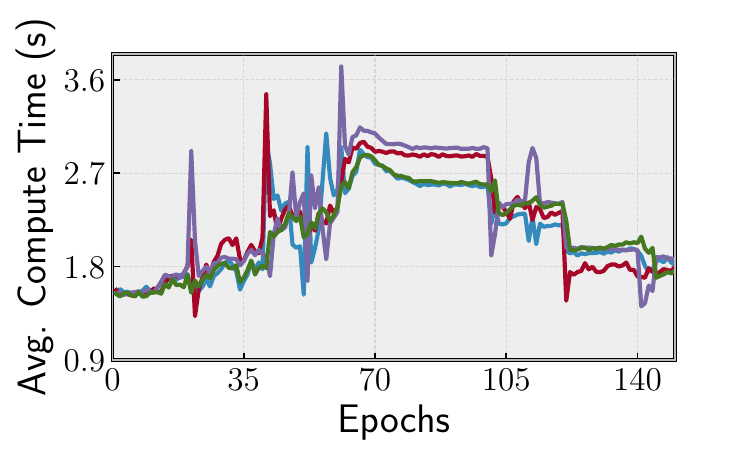}
    \caption{AlexNet compute-times}
    \label{fig:dynbatBSPtimeAlex}
  \end{subfigure}
  \hfill
  \begin{subfigure}{0.22\textwidth}
    \includegraphics[width=1.1\linewidth]{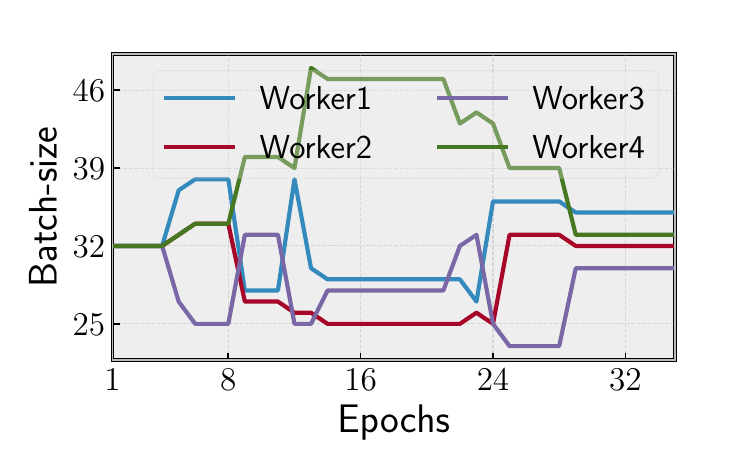}
    \caption{VGG11 batch-sizes}
    \label{fig:dynbatBSPbszVGG}
  \end{subfigure}
  \hfill
  \begin{subfigure}{0.22\textwidth}
    \includegraphics[width=1.1\linewidth]{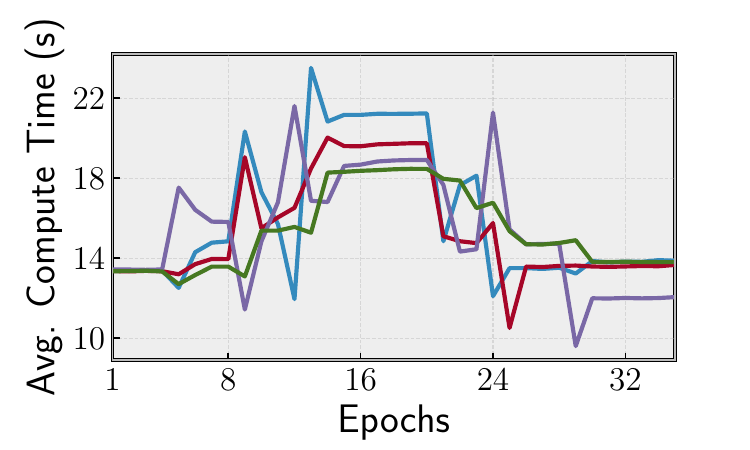}
    \caption{VGG11 compute-times}
    \label{fig:dynbatBSPtimeVGG}
  \end{subfigure}
  
  \caption{\texttt{OmniLearn} BSP training under \emph{dynamic} heterogeneity with $b_{min}=4$, $b_{max}=96$ and $\triangle b_{k}=10\%$ .}
  \label{fig:dynamicHLdeadbandBSP}
\end{figure*}

The batch adjustments and average computation times in BSP-based \texttt{OmniLearn} under dynamic heterogeneity is showed in Figure~(\ref{fig:dynamicHLdeadbandBSP}).
Across 4 workers, we set ($b_{min}$, $b_{max}$) to (4, 96), $\triangle b_{k}=0.1$ or 10\%, and impose a global batch-size constraint (of $32\times 4=128$).
For ResNet18 in Figures~(\ref{fig:dynbatBSPbszRes18}) and (\ref{fig:dynbatBSPtimeRes18}), the cluster is configured with \textit{HL1} between epochs 0-4 and thus, uses the same mini-batch 32 on every worker in that phase (uniform-batching).
Although we switch to \textit{HL2} at epoch 5, the workers still train with the previously chosen batches.
Thus, we observe a rapid variance in worker computation times where some containers get faster and others get much slower.
Worker \#3 has the least compute resources available to it (from Table~(\ref{table:computecoreHLs})) and the slowest in the cluster to compute its updates.
The latest compute times are collected over the epoch, and proportional-control adjustment (Equation~(\ref{eqn:dynpropctrl}) and (\ref{eqn:dynbszGlobalBFixed})) is applied post epoch 5 and prior epoch 6.
Hence, we see the average compute times across the cluster is reduced and stabilizes for \textit{HL2} between epochs 5-9.
Heterogeneity is then scaled from \textit{HL2} to \textit{HL4} at epoch 10, as evident from the abrupt change in computation time for that phase.
\texttt{OmniLearn} is able to efficaciously detect variances in worker performance and adjusts computation to reach steady-state equilibrium.
For ResNet50 in Figure~(\ref{fig:dynbatBSPtimeRes50}), we see that although PID-controller stabilizes computation times between epochs 50-150, the per-worker average compute times have noticeable variance.
This is because workers \#3 and \#4 with the least and most resources respectively have already saturated to the smallest and largest allowable mini-batch, i.e., $b_{min}$ and $b_{max}$ (Figure~(\ref{fig:dynbatBSPbszRes50})).
The batch-size of worker \#3 cannot be reduced any further, and \#4 cannot be increased in order to bring down the average cluster computation time.
In this case, performance is limited by the parameters chosen with \texttt{OmniLearn}'s configuration (specifically the min-max bounds).
For AlexNet and VGG11, appropriate batch-adjustments are made soon after \textit{HL} changes, so the effective compute times are soon equalized among workers.
\emph{Compared to uniform-batching that runs a fixed batch-size across all \textit{HL}s, \texttt{OmniLearn}'s dynamic-batching reduces TTA by 27.9\% in ResNet18, 22.72\% in ResNet50, 29.3\% in AlexNet and 12.8\% in VGG11.}.
Currently, PID-control adjustment is triggered at the end of an epoch.
However, we can tweak it to a finer granularity as well, say after every few iterations.

\begin{figure*}[h]
  \centering
  \begin{subfigure}{0.22\textwidth}
    \includegraphics[width=1.1\linewidth]{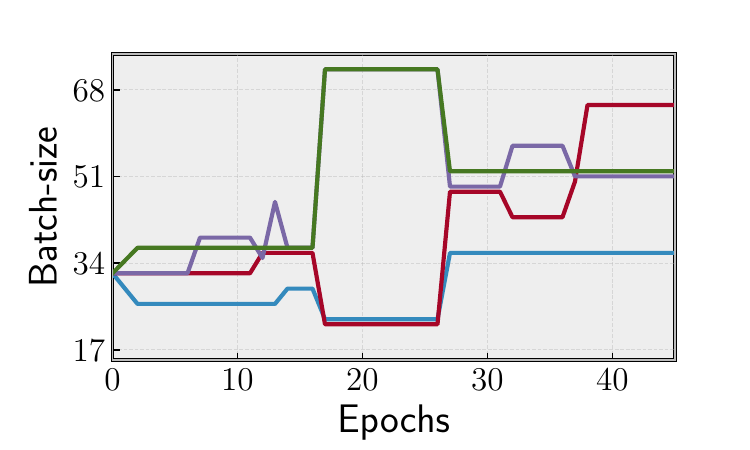}
    \caption{ResNet18 batch-sizes}
    \label{fig:dynbatASPbszRes18}
  \end{subfigure}
  \hfill
  \begin{subfigure}{0.22\textwidth}
    \centering
    \includegraphics[width=1.1\linewidth]{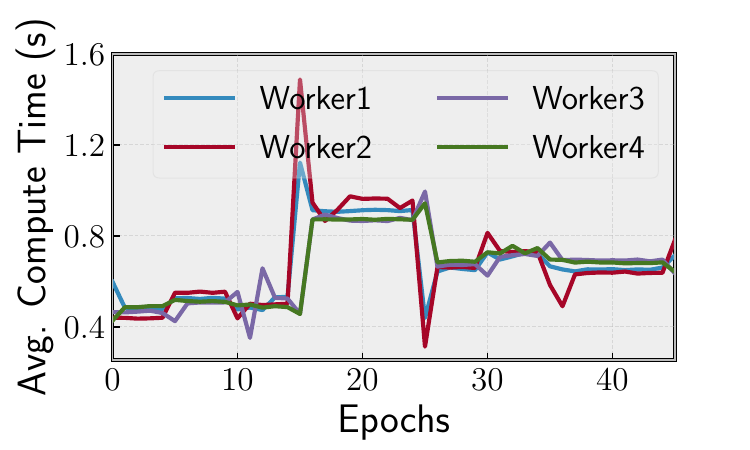}
    \caption{ResNet18 compute-times}
    \label{fig:dynbatASPtimeRes18}
  \end{subfigure}
  \hfill
  \begin{subfigure}{0.22\textwidth}
    \includegraphics[width=1.1\linewidth]{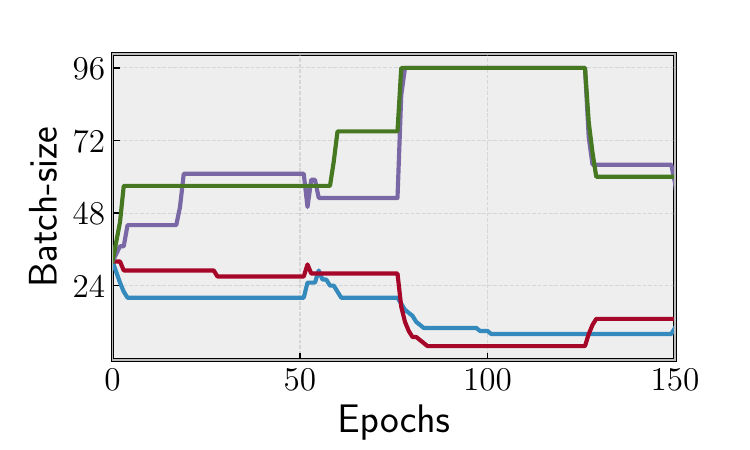}
    \caption{ResNet50 batch-sizes}
    \label{fig:dynbatASPbszRes50}
  \end{subfigure}
  \hfill
  \begin{subfigure}{0.22\textwidth}
    \includegraphics[width=1.1\linewidth]{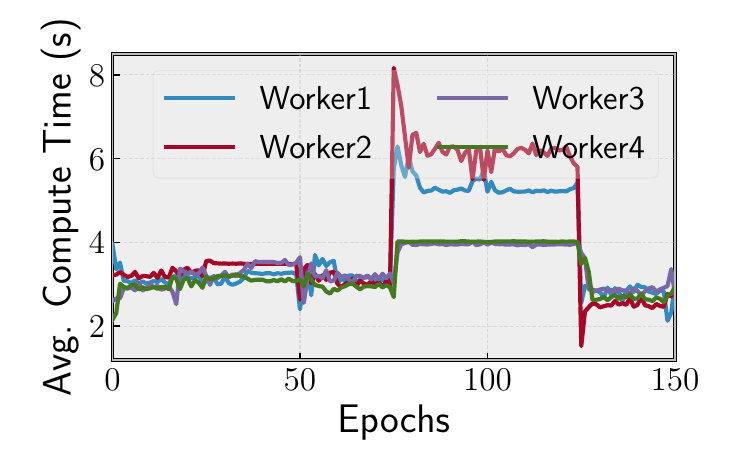}
    \caption{ResNet50 compute-times}
    \label{fig:dynbatASPtimeRes50}
  \end{subfigure}
  
  \vspace{1em}

  \begin{subfigure}{0.22\textwidth}
    \includegraphics[width=1.1\linewidth]{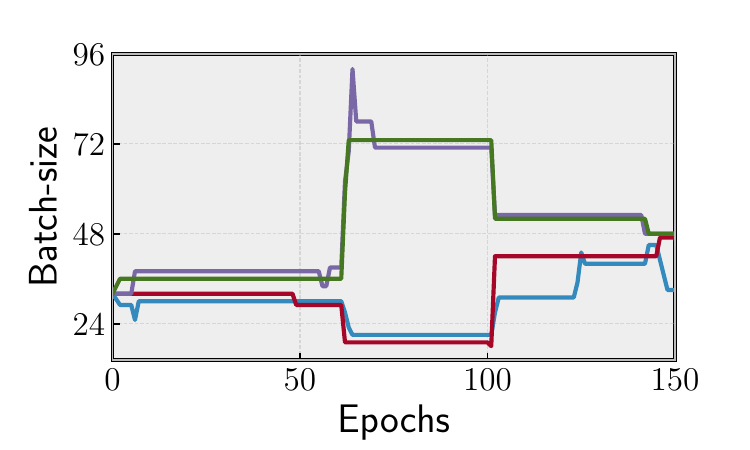}
    \caption{AlexNet batch-sizes}
    \label{fig:dynbatASPbszAlex}
  \end{subfigure}
  \hfill
  \begin{subfigure}{0.22\textwidth}
    \centering
    \includegraphics[width=1.1\linewidth]{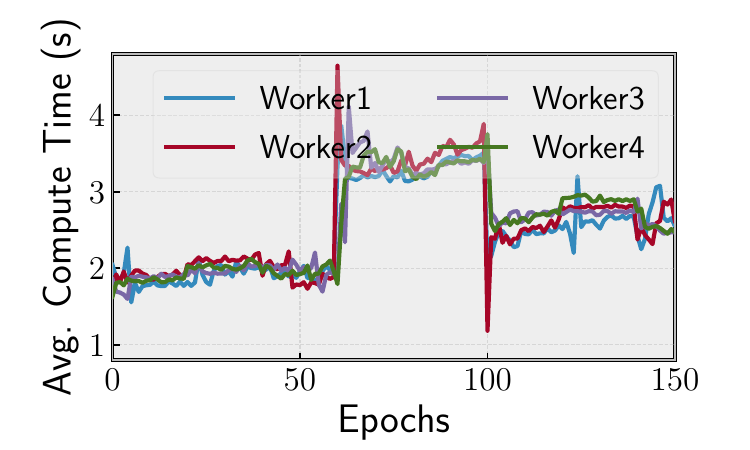}
    \caption{AlexNet compute-times}
    \label{fig:dynbatASPtimeAlex}
  \end{subfigure}
  \hfill
  \begin{subfigure}{0.22\textwidth}
    \includegraphics[width=1.1\linewidth]{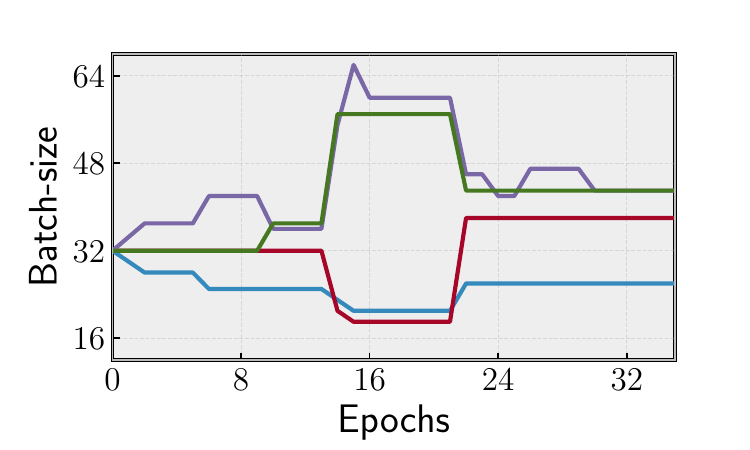}
    \caption{VGG11 batch-sizes}
    \label{fig:dynbatASPbszVGG}
  \end{subfigure}
  \hfill
  \begin{subfigure}{0.22\textwidth}
    \includegraphics[width=1.1\linewidth]{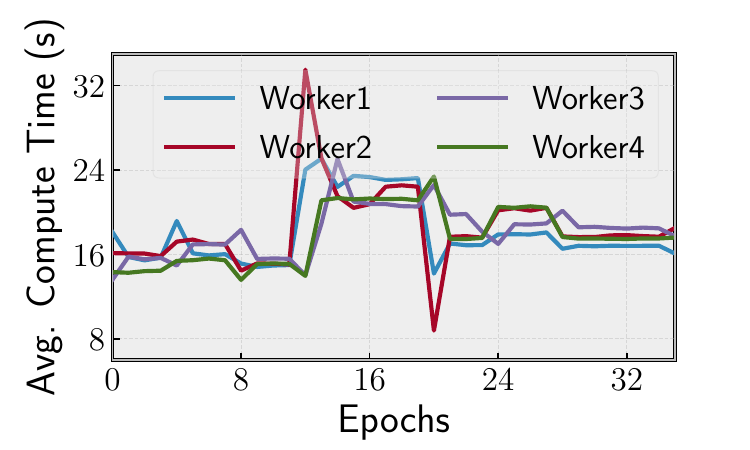}
    \caption{VGG11 compute-times}
    \label{fig:dynbatASPtimeVGG}
  \end{subfigure}
  
  \caption{\texttt{OmniLearn} ASP training under \emph{dynamic} heterogeneity with $b_{min}=4$, $b_{max}=96$  and $\triangle b_{k}=10\%$.}
  \label{fig:dynamicHLdeadbandASP}
\end{figure*}

Similar to BSP, we perform proportional-controlled adjustments in ASP under dynamic heterogeneity.
We keep the same configuration, $b_{min}=4$, $b_{max}=96$ and $\triangle b_{k}=0.1$.
To keep cumulative resources fixed between decentralized BSP and centralized ASP, the total cores allocated to workers are generally lower in ASP compared to BSP.
This is because ASP allocates additional resources to deploying a PS container.
Using worker allocations from Table~(\ref{table:computecoreHLs}), we vary heterogeneity for each model at intervals shown in Figure~(\ref{fig:dynheterogensimulation}).
With an initial worker batch-size of 32, batch-sizes are adjusted based on their respective computation times logged over the epochs.
One important thing to note is that the averaged estimates of per-worker compute times measured over the iterations in ASP may not be as accurate as BSP.
This is because BSP has a synchronization barrier and each worker performs its computation and logs its compute time at \emph{every} iteration.
However, in ASP training, the frequency of updates performed by workers is not the same; containers with more resources perform a larger fraction of updates on the PS than those with fewer resources.
The smaller workers can thus have higher variance in the measured compute times.
Still, we see in Figure~(\ref{fig:dynamicHLdeadbandASP}) that ASP-based \texttt{OmniLearn} is able to successfully adjust its batches proportional to worker compute times.
Akin to BSP, workers with more resources tend to have a lower computation cost and thus, afford a larger batch-size.
For e.g., ResNet18 at \textit{HL8} between epochs 15-24 has more resources on worker \#4 compared to other workers.
\texttt{OmniLearn} thus applied a larger batch-size on this worker is order to equalize compute times among workers to mitigate staleness.
ASP training does not impose a global batch-size rule; however a local limit of $b_{min}$ and $b_{max}$ is still applied.

\begin{figure}
	\centering
	\begin{subfigure}{0.22\textwidth}
		\includegraphics[width=1.1\linewidth]{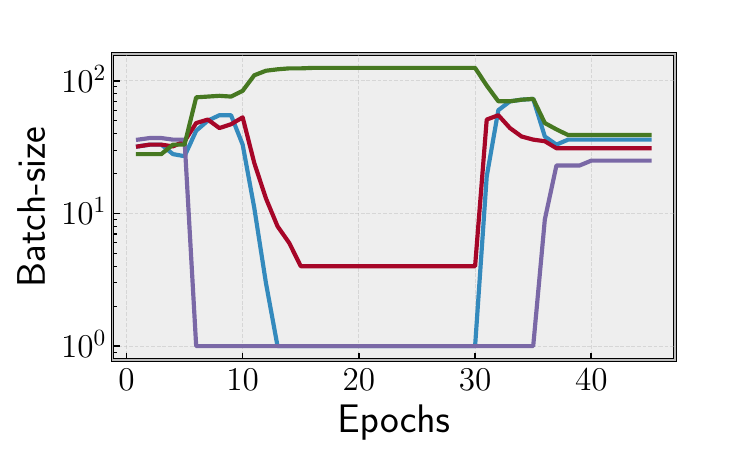}
   	 	\caption{GPT-2 batch-sizes}
   	 	\label{fig:gpt2dynamicHLbsz}
	\end{subfigure}
	\hspace{1.4em}
	\begin{subfigure}{0.22\textwidth}
		\centering
		\includegraphics[width=1.1\linewidth]{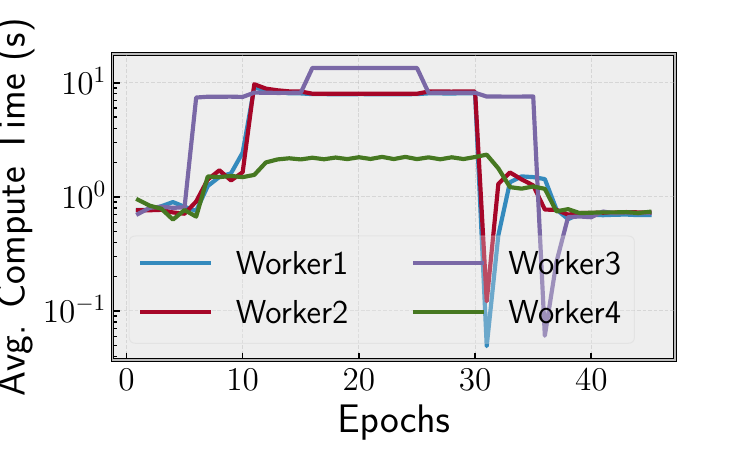}
		\caption{GPT-2 compute-times}
		\label{fig:gpt2dynamicHLcomputetime}
  	\end{subfigure}
  	
  	\caption{\texttt{OmniLearn} under dynamic heterogeneity for GPT-2 LLM training with $b_{min}=1$, $b_{max}=124$ and $\triangle b_{k}=5\%$.}
	\label{fig:gpt2omnilearntrain}
\end{figure}

\emph{Performance over large language models (LLM)}: Figure~(\ref{fig:gpt2omnilearntrain}) shows \texttt{OmniLearn} with GPT-2 (generative pre-training transformer) model \cite{bb47} trained over Shakespeare dataset under BSP as heterogeneity is emulated similar to ResNet18 in Figure~(\ref{fig:res18vggdynheterosim}).
We set initial batch-size to 32, $b_{min}$ to 1, $b_{max}$ to 124 and $\triangle b_{k}$ to 5\% respectively.
We see that training GPT-2 with \texttt{OmniLearn} actively adapts worker batches in response to dynamic heterogeneity.
At the \textit{HL8} phase between epochs 15-25, slow workers floor to $b_{min}$ while the fast worker ceils to $b_{max}$.
As a result, batches could not be optimized any further under extreme heterogeneity conditions.
In spite of this, performance is significantly improved over uniform-batching at high heterogeneity.
At \textit{HL8}, the difference in computation times between the slowest and fastest worker \emph{with} \texttt{OmniLearn} was under 12 seconds, while that of uniform-batching \emph{without} \texttt{OmniLearn} differed by over 22.7 seconds in GPT-2 training.
During the initial and end phases with \textit{HL1}, workers have identical resources so compute cost as well as the allocated batch-sizes are similar.

\subsection{Challenges and Limitations of \texttt{OmniLearn}}

\texttt{OmniLearn} improves computation cost across workers by proportionally partitioning the global batch-size.
Thus, high-resource workers process larger batches than others, and vice versa.
Training attention-based large language models over highly heterogeneous clusters \texttt{OmniLearn} can result in excessively large activation memory requirements as it rises proportionally to the batch-size, sequence length, number of transformer layers and hidden dimensions \cite{b35}.
Despite high-compute capability, training jobs on such nodes can fail from insufficient memory.
Thus, the batch-size bound $b_{max}$ should be chosen carefully for language models over a given cluster configuration, and determined prior-training as shown in \S \ref{subsubsec:hetmemOmnilearn}.
Even with dense and convolutional networks, choosing the right bounds ($b_{min}$, $b_{max}$) requires careful tuning as it affects overall training performance.
For e.g., ResNet50 in Figures~(\ref{fig:dynbatBSPbszRes50}) and (\ref{fig:dynbatBSPtimeRes50}) has high variance across worker compute times because the smallest and largest worker already hit the batch-size bounds.
Thus in extreme heterogeneity, we cannot fully remove stragglers if the ideal batch sizes are beyond these bounds. 
In this case, using a small $b_{min}$ and a larger $b_{max}$ could further accelerate training.

Another challenge in \texttt{OmniLearn} is with ASP under extreme heterogeneity settings where the estimated compute time for throughput estimation on slow workers may be inaccurate due to high variance.
This is because the average computation time logged over each batch adjustment in \texttt{OmniLearn} is measured for relatively fewer iterations on slow workers which may have high variance over limited samples.
Lastly, although \texttt{OmniLearn} effectively mitigates stragglers in BSP and staleness in ASP, it does not pick which of the two communication models is ideal for a given model and training configuration. 
This is because ASP and BSP training have different convergence-rates for a fixed batch-size and training iterations.
As a rule of thumb, BSP achieves better statistical performance, while ASP has a better parallel performance.
\section{Conclusion}

Heterogeneity is pervasive across the edge, shared clusters and cloud, so its crucial to develop systems and techniques for deploying neural networks in heterogeneous clusters.
We emulate varying degrees of heterogeneity with `\textit{Heterogeneity-level}' to account for the inherent variance in computational capability across nodes, and show how distributed learning suffers from stragglers in BSP and staleness in ASP training.
Although variable-batching slightly degrades the statistical performance in BSP, it improves parallel performance and reduces training time by 14-85\% over uniform-batching in highly heterogeneous settings.
To attenuate throughput estimation errors for more accurate workload distribution, \texttt{OmniLearn} uses PID controller-based batch adjustment mechanism that improves throughput and reduces training time.
Additional optimizations like weighted aggregation and learning-rate scaling improve accuracy by up to 10.21\% and 6.9\% respectively.
We also show how \texttt{OmniLearn}'s dynamic-batching is able to adapt to resource transiency in an online manner and improve performance over shared and heterogeneous clusters.
Our extensive empirical evaluation shows an average training speedup of 26\% while attaining comparable convergence targets.

\BlankLine
\BlankLine

\noindent \textbf{Acknowledgments:} Authors thank the reviewers for their feedback, and the department of Intelligent Systems Engineering at Indiana University Bloomington for providing the necessary computing infrastructure.
This work is supported in part by the National Science Foundation (NSF) grant OAC-2112606. 
Any opinions, findings, and conclusions or recommendations expressed in this material are those of the author(s) and do not necessarily reflect the views of the NSF.

\vspace{-3em}

\begin{IEEEbiography}[{\includegraphics[width=1.in,height=1.25in,clip,keepaspectratio]{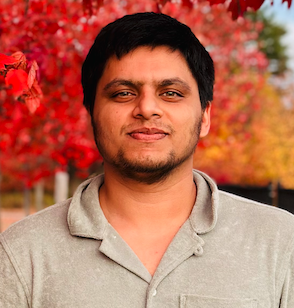}}]{Sahil Tyagi}
is a Ph.D. in Intelligent Systems Engineering (ISE) from the Luddy School of Informatics, Computing and Engineering at Indiana University Bloomington (IUB).
His current research focuses on distributed ML systems, cloud and high-performance computing, deep learning and federated training.
He can be contacted at sahilt.tyagi@gmail.com.
\end{IEEEbiography}

\vspace{-4em}

\begin{IEEEbiography}[{\includegraphics[width=1.in,height=1.25in,clip,keepaspectratio]{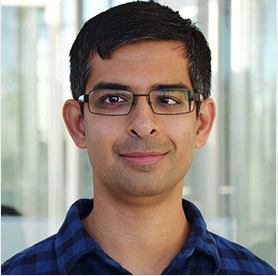}}]{Prateek Sharma}
is an Associate Professor in the ISE department at IUB. 
He has a Ph.D. in Computer Science from the University of Massachusetts Amherst
His current research focuses on Cloud Computing, Operating Systems and Virtualization. 
Sharma received his masters degree in Computer Science from Indian Institute of Technology, Bombay. 
Contact him at prateeks@iu.edu.
\end{IEEEbiography}


\begin{thebibliography}{1}
	\bibitem{b1} Li, M., Andersen, D.G., Park, J.W., Smola, A., Ahmed, A., Josifovski, V., Long, J., Shekita, E.J., and Su, B. (2014). Scaling Distributed Machine Learning with the Parameter Server. USENIX Symposium on Operating Systems Design and Implementation (OSDI), 2014.
	\bibitem{b2} Ang, K.H., Chong, G.C., and Li, Y. (2005). PID Control System Analysis, Design, and Technology. IEEE Transactions on Control Systems Technology, 13, 559-576.
	\bibitem{b3} Bottou, L. (2010). Large-Scale Machine Learning with Stochastic Gradient Descent. International Conference on Computational Statistics (COMPSTAT).
	\bibitem{b4} Thakur, R., Rabenseifner, R., and Gropp, W. (2005). Optimization of Collective Communication Operations in MPICH. The International Journal of High Performance Computing Applications (HPCA), 19, 49-66.
	\bibitem{b6} Agarwal, S., Wang, H., Venkataraman, S., and Papailiopoulos, D. (2021). On the Utility of Gradient Compression in Distributed Training Systems. ArXiv, abs/2103.00543.
	\bibitem{b7} McMahan, H. B., Moore, E., Ramage, D., Hampson, S., and Arcas, B. a. Y. (2017). Communication-Efficient Learning of Deep Networks from Decentralized Data. International Conference on Artificial Intelligence and Statistics (AISTATS), 1273–1282.
	\bibitem{bb78} Li, S., Zhao, Y., Varma, R., Salpekar, O., Noordhuis, P., Li, T., Paszke, A., Smith, J., Vaughan, B., Damania, P., and Chintala, S. (2020). PyTorch Distributed. Proceedings of the VLDB Endowment, 13, 3005-3018.
	\bibitem{b8} Jiang, Y., Zhu, Y., Lan, C., Yi, B., Cui, Y., and Guo, C. (2020). A Unified Architecture for Accelerating Distributed {DNN} Training in Heterogeneous GPU/CPU clusters. Operating Systems Design and Implementation (OSDI), 463–479.
	\bibitem{b10} Tyagi, S., and Swany, M. (2023). Accelerating Distributed ML Training via Selective Synchronization. 2023 IEEE International Conference on Cluster Computing (CLUSTER), 1-12.
	\bibitem{b11} Blot, M., Picard, D., Cord, M., and Thome, N. (2016). Gossip Training for Deep Learning. ArXiv, abs/1611.09726.
	\bibitem{b12} Lian, X., Zhang, W., Zhang, C., and Liu, J. (2017). Asynchronous Decentralized Parallel Stochastic Gradient Descent. ArXiv, abs/1710.06952.
	\bibitem{b13} Li, S., Walls, R.J., and Guo, T. (2020). Characterizing and Modeling Distributed Training with Transient Cloud GPU Servers. 2020 IEEE 40th International Conference on Distributed Computing Systems (ICDCS).
	\bibitem{bb1314} “EC2 Burstable Instances,” https://aws.amazon.com/blogs/aws/low-cost-burstable-ec2-instances, 2014.
	\bibitem{b14} Wang, C., Urgaonkar, B., Gupta, A., Kesidis, G., and Liang, Q. (2017). Exploiting Spot and Burstable Instances for Improving the Cost-efficacy of In-Memory Caches on the Public Cloud. Proceedings of the Twelfth European Conference on Computer Systems (EuroSys).
	\bibitem{b17} Tyagi, S., and Swany, M. (2023). GraVAC: Adaptive Compression for Communication-Efficient Distributed DL Training. 2023 IEEE 16th International Conference on Cloud Computing (CLOUD), 319-329.
	\bibitem{b18} Recht, B., Ré, C., Wright, S.J., and Niu, F. (2011). Hogwild!: A Lock-Free Approach to Parallelizing Stochastic Gradient Descent. Neural Information Processing Systems (NIPS).
	\bibitem{b19} McCandlish, S., Kaplan, J., Amodei, D., and Team, O.D. (2018). An Empirical Model of Large-Batch Training. ArXiv, abs/1812.06162.
	\bibitem{b20} Yao, Z., Gholami, A., Keutzer, K., and Mahoney, M.W. (2018). Large batch size Training of Neural Networks with Adversarial Training and Second-Order Information. ArXiv, abs/1810.01021.
	\bibitem{b21} Ferdinand, N.S., Al-Lawati, H., Draper, S.C., and Nokleby, M.S. (2020). Anytime Minibatch: Exploiting Stragglers in Online Distributed Optimization. ArXiv, abs/2006.05752.
	\bibitem{b22} Keskar, N.S., Mudigere, D., Nocedal, J., Smelyanskiy, M., and Tang, P.T. (2016). On Large-Batch Training for Deep Learning: Generalization Gap and Sharp Minima. ArXiv, abs/1609.04836.
	\bibitem{b23} Hoffer, E., Hubara, I., and Soudry, D. (2017). Train longer, generalize better: closing the Generalization Gap in Large Batch Training of neural networks. ArXiv, abs/1705.08741.
	\bibitem{b24} Goyal, P., Dollár, P., Girshick, R.B., Noordhuis, P., Wesolowski, L., Kyrola, A., Tulloch, A., Jia, Y., and He, K. (2017). Accurate, Large Minibatch SGD: Training ImageNet in 1 Hour. ArXiv, abs/1706.02677.
	\bibitem{b16} Tyagi, S., and Sharma, P. (2023). Scavenger: A Cloud Service For Optimizing Cost and Performance of ML Training. 2023 IEEE/ACM 23rd International Symposium on Cluster, Cloud and Internet Computing (CCGrid), 403-413.
	\bibitem{b25} Krizhevsky, A. (2014). One weird trick for parallelizing Convolutional Neural Networks. ArXiv, abs/1404.5997.
	\bibitem{b26} He, K., Zhang, X., Ren, S., and Sun, J. (2015). Deep Residual Learning for Image Recognition. 2016 IEEE Conference on Computer Vision and Pattern Recognition (CVPR), 770-778.
	\bibitem{b27} Bossard, L., Guillaumin, M., and Gool, L.V. (2014). Food-101 - Mining Discriminative Components with Random Forests. European Conference on Computer Vision (ECCV).
	\bibitem{b28} Krizhevsky, A., Sutskever, I., and Hinton, G.E. (2012). ImageNet Classification with Deep Convolutional Neural Networks. Communications of the ACM, 60, 84 - 90.
	\bibitem{b29} Fei-Fei, L., Fergus, R., and Perona, P. (2004). Learning Generative Visual Models from Few Training Examples: An Incremental Bayesian Approach Tested on 101 Object Categories. 2004 Conference on Computer Vision and Pattern Recognition (CVPR) Workshop, 178-178.
	\bibitem{b30} Griffin, G., Holub, A., and Perona, P. (2007). Caltech-256 Object Category Dataset.
	\bibitem{b31} Simonyan, K., and Zisserman, A. (2014). Very Deep Convolutional Networks for Large-Scale Image Recognition. CoRR, abs/1409.1556.
	\bibitem{b32} Zhou, B., Lapedriza, À., Khosla, A., Oliva, A., and Torralba, A. (2018). Places: A 10 Million Image Database for Scene Recognition. IEEE Transactions on Pattern Analysis and Machine Intelligence, 40.
	\bibitem{b33} Docker https://docs.docker.com/config/containers/resource\_constraints/.
	\bibitem{b34} Smith, S.L., Kindermans, P., and Le, Q.V. (2017). Don't Decay the Learning Rate, Increase the Batch Size. ArXiv, abs/1711.00489.
	\bibitem{b35} Rajbhandari, S., Rasley, J., Ruwase, O., and He, Y. (2019). ZeRO: Memory optimizations Toward Training Trillion Parameter Models. SC20: International Conference for High Performance Computing, Networking, Storage and Analysis (SC'20), 1-16.
	\bibitem{b37} Tyagi, S., and Swany, M. (2023). Flexible Communication for Optimal Distributed Learning over Unpredictable Networks. 2023 IEEE International Conference on Big Data (BigData), 925-935.
	\bibitem{b38} Zhang, S., Choromańska, A., and LeCun, Y. (2014). Deep learning with Elastic Averaging SGD. Neural Information Processing Systems (NIPS).
	\bibitem{bb39} Tyagi, S., and Swany, M. (2022). ScaDLES: Scalable Deep Learning over Streaming data at the Edge. 2022 IEEE International Conference on Big Data (Big Data), 2113-2122.
	\bibitem{bb40} Zhou, Q., Guo, S., Qu, Z., Li, P., Li, L., Guo, M., and Wang, K. (2021). Petrel: Heterogeneity-Aware Distributed Deep Learning Via Hybrid Synchronization. IEEE Transactions on Parallel and Distributed Systems (TPDS), 32, 1030-1043.
	\bibitem{bb41} Jiang, J., Cui, B., Zhang, C., and Yu, L. (2017). Heterogeneity-aware Distributed Parameter Servers. Proceedings of the 2017 ACM International Conference on Management of Data.
	\bibitem{bb42} Luo, Q., Lin, J., Zhuo, Y., and Qian, X. (2019). Hop: Heterogeneity-aware Decentralized Training. Proceedings of 24th International Conference on Architectural Support for Programming Languages and Operating Systems (ASPLOS).
	\bibitem{bb4275} Tyagi, Sahil. "Towards Building Efficient Computation and Communication Models for Deep Learning Systems" \href{https://www.proquest.com/openview/4095235239d9328ddc2e47d5d6ccb337/}{ProQuest}, 2025.
	\bibitem{bb43} Miao, X., Nie, X., Shao, Y., Yang, Z., Jiang, J., Ma, L., and Cui, B. (2021). Heterogeneity-Aware Distributed Machine Learning Training via Partial Reduce. Proceedings of the 2021 International Conference on Management of Data.
	\bibitem{bb44} Li, S., Ben-Nun, T., Girolamo, S.D., Alistarh, D., and Hoefler, T. (2019). Taming unbalanced training workloads in deep learning with partial collective operations. Proceedings of the 25th ACM SIGPLAN Symposium on Principles and Practice of Parallel Programming (PPoPP).
	\bibitem{bb45} Chen, J., Monga, R., Bengio, S., and Józefowicz, R. (2016). Revisiting Distributed Synchronous SGD. ArXiv, abs/1604.00981.
	\bibitem{bb46} Luo, Q., He, J., Zhuo, Y., and Qian, X. (2020). Prague: High-Performance Heterogeneity-Aware Asynchronous Decentralized Training. Proceedings of 25th International Conference on Architectural Support for Programming Languages and Operating Systems (ASPLOS).
	\bibitem{bb47} Radford, A., and Narasimhan, K. (2018). Improving Language Understanding by Generative Pre-Training.
	\bibitem{bb48} Zhang, H., Zheng, Z., Xu, S., Dai, W., Ho, Q., Liang, X., Hu, Z., Wei, J., Xie, P., and Xing, E.P. (2017). Poseidon: An Efficient Communication Architecture for Distributed Deep Learning on GPU Clusters. ArXiv, abs/1706.03292.
	\bibitem{bb49} Ho, Q., Cipar, J., Cui, H., Lee, S., Kim, J.K., Gibbons, P.B., Gibson, G.A., Ganger, G.R., and Xing, E.P. (2013). More Effective Distributed ML via a Stale Synchronous Parallel Parameter Server. Advances in neural information processing systems (NIPS), 2013, 1223-1231.
	\bibitem{bb50} Tyagi, Sahil and Sharma, Prateek. “Taming Resource Heterogeneity In Distributed ML Training With Dynamic Batching.” 2020 IEEE International Conference on Autonomic Computing and Self-Organizing Systems (ACSOS): 188-194.
\end{thebibliography}
\end{document}